\documentclass[preprint,12pt,authoryear]{elsarticle}

\usepackage{amsmath}
\usepackage{amsfonts}
\usepackage{booktabs} 
\usepackage{caption} 
\usepackage{subcaption} 
\usepackage{graphicx}
\usepackage{pgfplots}
\usepackage[all]{nowidow}
 \usepackage{verbatim}
\usepackage[utf8]{inputenc}
\usepackage[linesnumbered,ruled,vlined]{algorithm2e}
\usepackage{tikz}
\usetikzlibrary{er,positioning,bayesnet}
\usepackage{multicol}
\usepackage{listings}
\usepackage{hyperref}
\usepackage{multirow}
\usepackage{float}
\usepackage[inline]{enumitem} 
\usepackage{placeins}
\usepackage[a4paper]{geometry}

\journal{Preprint submitted for review}

\pgfplotsset{compat=1.18}
\begin{document}

\begin{frontmatter}



\title{Vacuum Spiker: A Spiking Neural Network-Based Model for Efficient Anomaly Detection in Time Series}

\author[itcl]{I. X. V\'azquez}
\author[itcl]{J. Sedano}
\author[birmingham]{M. Afzal}
\author[dasci]{A. M. Garc\'ia-Vico}

\address[itcl]{ITCL Technology Center, L\'opez Bravo St. 70, 01001 Burgos,
Castilla y Le\'on, Spain}

\address[birmingham]{Faculty of Computing, Engineering and the Built Environment, Birmingham City University, Birmingham, United Kingdom}

\address[dasci]{Andalusian Research Institute in Data Science and Computational Intelligence (DaSCI), Campus Las Lagunillas, s/n 23071, Universidad de Ja\'en, Ja\'en,
Andaluc\'ia, Spain}

%
\begin{abstract}

Anomaly detection is a key task across domains such as industry, healthcare, and cybersecurity. Many real-world anomaly detection problems involve analyzing multiple features over time, making time series analysis a natural approach for such problems. While deep learning models have achieved strong performance in this field, their trend to exhibit high energy consumption limits their deployment in resource-constrained environments such as IoT devices, edge computing platforms, and wearables. To address this challenge, this paper introduces the \textit{Vacuum Spiker algorithm}, a novel Spiking Neural Network-based method for anomaly detection in time series. It incorporates a new detection criterion that relies on global changes in neural activity rather than reconstruction or prediction error. It is trained using Spike Time-Dependent Plasticity in a novel way, intended to induce changes in neural activity when anomalies occur. A new efficient encoding scheme is also proposed, which discretizes the input space into non-overlapping intervals, assigning each to a single neuron. This strategy encodes information with a single spike per time step, improving energy efficiency compared to conventional encoding methods. Experimental results on publicly available datasets show that the proposed algorithm achieves competitive performance while significantly reducing energy consumption, compared to a wide set of deep learning and machine learning baselines. Furthermore, its practical utility is validated in a real-world case study, where the model successfully identifies power curtailment events in a solar inverter. These results highlight its potential for sustainable and efficient anomaly detection.


\end{abstract}


\end{frontmatter}

\section{Introduction}

In recent years, deep learning algorithms have been applied in many fields, such as computer vision \citep{Voulodimos2018Deep}, language and image generation \citep{Touvron2023LLaMAOA,Alemohammad2024SelfImprovingDM}, or sentiment analysis \citep{Zhang2018Deep}, achieving impressive results. One of these fields is anomaly detection \citep{9005687}. Anomaly detection algorithms address the problem of identifying cases that deviate from usual behaviour. 

Anomaly detection problems are critical in many domains. For example, in industry, those kind of algorithms can assist in quality control during production \citep{LIU2021106324,Zope2019Anomaly}, detect problems in machines \citep{Pittino2020Automatic}, or monitor air quality \citep{Hu2018Detecting}, among others. In the healthcare domain, anomaly detection methods can identify potential illnesses \citep{10.1007/978-3-031-16452-1_4}, or assist in medical image analysis \citep{10.1007/978-3-031-16452-1_4}. In cybersecurity, those methods enable the detection of malicious behaviour in networks \citep{Atefi2016AnomalyDB}, or anomalies in bank account transactions \citep{Wang2020Egonet}. In financial analysis, anomaly detection methods have also been applied to stock market data \citep{Golmohammadi2015Time}. 

In general, anomaly detection problems involve the study of data collected from different sensors over time, so time series analysis arises as a natural approach for designing algorithms to address such a problem. For this reason, deep neural networks specialized in capturing the temporal dependencies that exist among these data, such as Long Short Term Memories (LSTM) \citep{Lee2023Anomaly}, Gated Recurrent Units (GRU) \citep{Lee2018CNN}, or Transformers \citep{Xu2021AnomalyTT} have been extensively applied to anomaly detection. However, deep learning algorithms have been often associated with high energy consumption \citep{Getzner2023AccuracyIN}, due to their need for high computational power to function effectively. This poses challenges for deployment in IoT environments \citep{10.1145/3578938} or on battery-powered edge devices \citep{Chen2019Deep}. This concern has spurred ongoing efforts toward greater efficiency, with methods such as quantization and pruning \citep{10.1145/3578938}, aimed at reducing models complexity. In fact, the growing awareness of the environmental and economic impact of AI models has led to the emergence of the \textit{Green AI} paradigm \citep{10.1145/3381831}, which advocates for energy-efficient techniques and hardware optimizations as a means of aligning AI development with sustainability goals.

One of the proposed approaches for developing more energy-efficient models than current deep learning ones is the use of Spiking Neural Networks (SNNs) \citep{MAASS19971659}. An SNN is a dynamic system that processes information through sparse, asynchronous, and binary signals referred to as \textit{spikes} \citep{Pfeiffer2018DeepLW}. In an SNN, Multiply-Accumulate operations (MAC) are only performed when voltage updates occur in neurons or when a spike crosses a connection. In contrast, a traditional ANN requires a MAC operation for each connection every time the network is applied. Therefore, since there tend to be more connections than neurons in a neural network, if the number of spikes in the SNN is kept low, this can result in more energy-efficient models than an ANN \citep{9522999}. 

In addition to their energy efficiency, SNNs show several key advantages for performing anomaly detection on time series. First, because of their dynamic behaviour, SNNs can adapt and evolve over time through precise activation timings \citep{Saunders2018STDPLO}, which makes them especially interesting for capturing complex temporal patterns. Second, they can potentially react quickly \citep{Bassler2022}, which could lead to earlier anomaly detection on time series ---something that can be critical in multiple domains, such as industry or cybersecurity. Third, classification or anomaly detection can be performed by monitoring the spiking activity of certain neurons over time, as shown in \citep{Diehl2015-vb}, which prevents the need for time windows over past data. This, in turn, can reduce the number of parameters involved in tuning and allows the development of lighter models.
This feature, in addition to boosting energy efficiency, is especially important for deployment at the edge, where computing systems are often resource-constrained, and the use of complex models could carry to system slowdown. 

In this paper, the \textit{Vacuum Spiker} algorithm is proposed. This is a Spiking Neural Network-based model designed to perform anomaly detection in time series data in a highly efficient way. Its main contributions can be summarized as follows:
\begin{itemize}

\item The spiking activity of hidden neurons is directly monitored, avoiding the need for reconstruction or prediction error calculations, enhancing the efficiency of the proposed model.

\item Spike Time-Dependent Plasticity (STDP) is applied in a new way that enforces the prevalence of either potentiation or depression events for a given connection, instead of focusing on the relative order of pre- and post-synaptic spikes. Thus, connections between layers can be forced to exhibit either a prevalent excitatory or inhibitory behaviour, which can be exploited to keep a lower activity in hidden neurons when normal data are presented, increasing it when patterns in input data differ from the learnt ones

\item The algorithm is fed through a new single-spike coding strategy, where each input datum is represented by a single spike. The proposed coding scheme does not require exposing input data across several time steps, which prevents the use of artificial time windows, as is the case in other coding schemes used in SNNs, like rate coding, temporal coding, or population coding. The proposed coding scheme enhances the efficiency of the Vacuum Spiker even further.

\end{itemize}

The rest of the article is organized as follows: In Section \ref{trasfondo}, the techniques and technologies used to develop the proposed method are detailed, and similar works to ours are discussed. In Section \ref{metodo}, the Vacuum Spiker algorithm is described in detail. In Section \ref{experimental}, the experimental design is described, followed by Section \ref{resultados}, where the obtained results are presented. In Section \ref{solar}, a real case application is discussed, and in Section \ref{conclusion}, we present our conclusions on the results obtained.

The source code used to generate the results presented in this work is avalilable at \url{https://github.com/iago-creator/Vacuum_Spiker_experimentation}.

\section{Background}
\label{trasfondo}

In recent years, the advancement of machine learning and deep learning models has transformed the field of anomaly detection. From classical machine learning approaches to deep learning architectures, these methods have shown a strong ability to model non-linear patterns and extract useful representations across diverse domains. More recently, SNNs have emerged as an alternative with the potential to provide significant advantages in energy efficiency. In this section, the background on these approaches is introduced, covering traditional statistical and machine learning methods, deep learning models, and SNNs.

\subsection{Anomaly Detection}

Anomaly detection refers to the identification of data instances that significantly deviate from the expected pattern within a dataset \citep{Chalapathy2019DeepLF}. It is a critical task across diverse domains, including manufacturing \citep{LIU2021106324,Pittino2020Automatic}, healthcare \citep{10.1007/978-3-031-16452-1_4,10.1007/978-3-030-59710-8_46}, and cybersecurity \citep{Atefi2016AnomalyDB}. 

Traditionally, a variety of approaches have been proposed to tackle this problem. The most conventional are statistical methods, which model the data distribution to identify outliers \citep{Madhuri2020StatisticalAT}. These methods are simple and computationally efficient but often rely on strong assumptions about the data and may be inadequate for complex or non-parametric distributions \citep{osei2024intelligent,yurchuk2023quantile}.

With the growing use of algorithmic approaches, including machine learning, more sophisticated methods have been introduced. These can be broadly categorized into the following three groups:

\begin{itemize}

\item \textit{Supervised Approaches:} Supervised methods apply machine learning or deep learning classifiers to distinguish between normal and anomalous instances \citep{10.1007/978-3-319-10422-5_21}, or among multiple anomaly types \citep{8706523}. Within the machine learning domain, Random Forests \citep{modi2025anomaly,10.1007/978-3-031-36822-6_29} and Support Vector Machines (SVMs) \citep{Mathar2020SupportVM} are commonly used. For instance, \citep{more2024hybrid} applied Random Forest to credit card fraud detection, while \citep{10180421} used Random Forest, SVMs, and Naive Bayes for intrusion detection systems. The application of Random Forest and SVMs for software fault detection was studied by \citep{Agarwal2024ENHANCINGFD}. Deep learning algorithms have also been used. In the context of deep learning, Convolutional Neural Networks (CNNs) combined with the You Only Look Once (YOLO) architecture have been employed to traffic accident detection \citep{9001731},whereas deep neural networks have been applied to denial of service (DoS) attack detection \citep{Ahmed2020DeepLF}. CNNs have also been used for anomaly detection in industrial quality control \citep{Ahmed2020DeepLF}.

These models can achieve high accuracy when trained on well-labelled data \citep{10180421}. However, due to the scarcity of anomalies and the dominance of normal instances, they often suffer from class imbalance issues \citep{Ramadhan2024PerformanceCO,RoblesDurazno2018ASE}.

\item \textit{Unsupervised Approaches:} To address the challenge of labelled data scarcity, unsupervised and semi-supervised methods have been developed. Unsupervised techniques do not require labelled data \citep{9892807} and include clustering algorithms such as K-Means \citep{Chong2021KmeansCA} or DBSCAN \citep{Deng2020DBSCANCA}. Other popular algorithms are k-Nearest Neighbours of Neighbours (k-NNN) \citep{Nizan2023kNNNNN}, and Isolation Forest \citep{Downey2024AnomalyDW}.

For example, \citep{Hairach2023AnomalyDI} employed K-Means and DBSCAN to detect anomalies in photovoltaic modules with high accuracy. These algorithms were also applied in payment fraud detection \citep{Arvalo2022IdentifyingCO}. Clustering algorithms were compared with Isolation Forest for cybersecurity \citep{Fernando2024EvaluationOT}, finding the latter more computationally efficient. k-Nearest Neighbours (k-NN) and Isolation Forest were used for anomaly detection in tea traceability \citep{Yang2022UnsupervisedOD}, and Isolation Forest was also applied in geological data \citep{Janjua2024BigDA}.

While unsupervised methods offer the advantage of operating without labelled data ---which is an important advantage in many real-world scenarios--- they may yield false positives, as outliers do not necessarily represent true anomalies \citep{Paradhi2024AnomalyDI}.

\item \textit{Semi-Supervised Approaches:} Semi-supervised methods are trained exclusively on normal data to identify deviations during inference \citep{Noto2012,Chalapathy2019DeepLF}, which makes them better suited for addressing the scarcity of anomalies compared to supervised classifiers. The most widely used machine learning algorithm in this category is the One-Class Support Vector Machine (OCSVM). For example, \citep{10394193} used OCSVM to detect anomalies in hydraulic systems, \citep{Vos2022VibrationbasedAD} captured mechanical faults via vibration analysis, and \citep{esmaeilzadeh2022abusefrauddetectionstreaming} applied it to fraud detection in streaming services. However, OCSVMs present the disadvantage of being computationally intensive, limiting their applicability to real-time or large-scale anomaly detection \citep{zhu2010anomalous,10.1007/978-3-030-10925-7_10,zhu2010anomalous}.

Within the deep learning domain, semi-supervised approaches are the most prevalent. A common strategy involves training a model to reconstruct its input, and then measuring the reconstruction error as an indicator of anomalous behaviour \citep{Zenati2018EfficientGA}. Autoencoders (AEs) \citep{Pawar2019AssessmentOA}, Variational Autoencoders (VAEs) \citep{Kumar2023Anomaly}, and Generative Adversarial Networks (GANs) \citep{Zenati2018EfficientGA} are frequently utilized. For example, \citep{Minhas2020SemisupervisedAD} applied AEs to industrial optical inspection, while \citep{Guo2019BaggingDA} introduced a bagging of AEs to detect anomalies in images and in spacecraft payloads. AEs were used for detecting anomalies in high-performance computing systems \citep{Borghesi2019ASA}, as well as to cybersecurity, where Bayesian variants have been explored \citep{CasajsSetin2022EvolutiveAB}.

VAEs have also been successfully applied in various domains. They were used to detect anomalies in dermatological images \citep{Lu2018AnomalyDF}, while \citep{Kumarage2018AnomalyDI} applied them to industrial software systems. \citep{Pol2019AnomalyDW} used VAEs to monitor the trigger system at the CERN Large Hadron Collider. In the case of GANs, \citep{HASHIMOTO2021873} employed them in semiconductor manufacturing, \citep{10826126} used them with urban sensor networks, and \citep{Kim2023AGA} applied them to detect anomalies in stock market prices.

However, deep learning semi-supervised methods present some disadvantages. AEs can sometimes reconstruct anomalies too well ---thus reducing detection performance \citep{Astrid2024ExploitingAW}. GANs are computationally expensive, and require complex hyperparameter tuning \citep{10804166,Shyju2023ATLASA}. To mitigate these issues, some methods incorporate a small set of labelled anomalies into the training process to enhance model performance. For example, \citep{Angiulli2023ReconstructionEA} proposed a method to increase the separation between normal and anomalous samples using labelled data in AE-like architectures. Mutual information and entropy between latent representations have been also explored to better distinguish between normal and anomalous data \citep{Huang2020ESADED}. Similarly, entropy-based methods have been leveraged to improve separation between those two kinds of data \citep{Ruff2019DeepSA}.

\end{itemize}
\subsection{Anomaly Detection in Time Series}

For time series data, specialized models have been adopted. The most common is the AutoRegressive Integrated Moving Average (ARIMA) model \citep{stram1986temporal}, a statistical approach originally designed for time series forecasting. When employed in anomaly detection, deviations between predictions and actual observations are used to flag anomalies \citep{6113293}. Applications include detecting anomalies in web service key performance indicators \citep{Shi2018AnomalyDF}, data streams \citep{Hasani2019SurveyAP}, IP traffic \citep{Pena2013AnomalyDU}, and network attacks \citep{Hulskamp2022EffectivenessAO}. While ARIMA performs well on time series processing, it struggles with complex or non-linear behaviours \citep{cmes2023045251}. Additionally, it can be inefficient for large-scale datasets \citep{Liu_Hoi_Zhao_Sun_2016}.

To address these limitations, deep learning approaches have been increasingly adopted for time series anomaly detection. Conceptually, the manner these methods are applied resembles the supervised and semi-supervised approaches discussed earlier. The most relevant difference is the incorporation of models designed for sequential data, such as Long Short-Term Memory (LSTM) networks, Gated Recurrent Units (GRUs), and Transformers. For example, \citep{Maru2020CollectiveAD} combined GANs with a Sequence-to-Sequence model to detect abnormal patterns in multivariate time series. A VAE combined with a bidirectional LSTM (Bi-LSTM) has also been applied to anomaly detection in electrocardiogram (ECG) data \citep{8679157}, while an LSTM-based VAE has been proposed for analysing sequences of health events \citep{9005687}.

Among these approaches, LSTM-based models are the most widely used. They have been applied to detect anomalies in different sectors, such as healthcare (e.g., arrhythmias \citep{OH2018278,YILDIRIM2019121}), industry (e.g., water pumps \citep{9613542}), ecology (e.g., river catchment water levels \citep{githinji2023ciira}), and cybersecurity \citep{Lee2023Anomaly}). Nonetheless, recent studies have applied CNNs for anomaly detection in sequential data, despite these models not being originally designed for such tasks \citep{Lee2018CNN,9925612}.

In deep learning-based anomaly detection, reconstruction error is common, but prediction error, the gap between the predicted and actual next step, is also often used in time-series analysis. That is the case of \citep{10.1007/978-3-319-93034-3_46}, which proposed a framework combining statistical and deep learning methods to detect anomalies in time series data; and \citep{8581424}, which introduced a CNN-based architecture that relies on prediction error as the anomaly criterion.

\subsection{Spiking Neural Networks}

An SNN is a kind of Artificial Neural Network that processes data through sparse, asynchronous binary signals referred to as spikes \citep{Pfeiffer2018DeepLW}. In an SNN, one MAC operation is performed each time a spike crosses a connection, and when voltage updates are produced in neurons. Meanwhile, the number of MAC operations in a traditional ANN mainly depends on the number of connections, which tends to be higher than the number of neurons, For this reason, if the number of spikes that are generated remains low, SNNs will be more energy efficient than traditional ANNs.

SNNs have been applied to different areas, including neuroscience \citep{Stimberg2020}, robotics \citep{Yamazaki2022SpikingNN} and computer vision \citep{Hopkins2018SpikingNN}. Due to their inherent structure, SNNs are particularly well-suited for processing temporal data. For instance, \citep{Iaboni2024EventBasedSN} used them to process input from event-based cameras. In the context of time series processing, they have also been utilized for tasks such as forecasting \citep{Lucas2024MethodologyBO} and classification \citep{Fang2020MultivariateTS}. For example, \citep{Reid2014Financial} used them to predict the financial market, and \citep{Sharma2010A}, for forecasting in electric markets. They have also been explored in anomaly detection, as in \citep{9291232}, where they were used to identify car hacking attempts.

\subsubsection{Coding}

In an SNN, a method to transform numerical inputs into spikes and outputs into the desired targets is needed to process information through spikes. Three steps have to be taken when using an SNN: first, numeric inputs have to be coded into spikes, then, the spikes are processed in the SNN, and finally, the outputs have to be decoded. The way inputs are coded can have a significant impact on the number of spikes that are generated \citep{8351295}, SNN latency \citep{10.3389/fnins.2021.638474}, and model performance \citep{Yarga2022Efficient}, so different coding strategies have been developed. Among them, some of the most used are rate coding \citep{7727355}, which is based on assigning higher spike frequency to higher input values; Time-to-First-Spike (TTFS) \citep{Zhang2019TDSNNFD}, where higher values signify earlier spikes; burst coding \citep{10.1145/3316781.3317822}, similar to TTFS, but where information is coded as a single burst of spikes instead of using only one spike, or phase coding \citep{KIM2018373}, where information is converted into binary representation, so that 1 is the generation of a spike. 

In time series analysis, population encoding \citep{Fang2020MultivariateTS}, where each input value activates a group of neurons to varying degrees —--allowing different neurons to respond more or less strongly depending on their tuning to the stimulus--- has also been applied. The degree of activation of each neuron is balanced through a complementary coding scheme, such as rate coding or temporal coding. Alternatively, direct encoding \citep{10191614}, where numeric input values are added directly to the membrane potentials of the input-layer neurons, has also been explored.

Most of the aforementioned coding strategies rely on artificial time windows to code each individual record, which may increase computational complexity and introduce latency ---factors that could affect their suitability for real-time anomaly detection in time series. In addition, several of them, like rate coding, or population coding, may require a higher number of spikes to perform the conversion, which might lead to increased energy consumption, posing challenges in production settings if energy supply is limited, thereby potentially constraining the deployment of the model on wearable or battery-powered devices.

\subsubsection{Leaky Integrate and Fire Neuron}

Different models of spiking neurons have been developed over time. Many of them are intended to mimic biological neurons, to better understand their behaviour in a neuroscientific context, such as the Hodkin and Huxley model \citep{Hodgkin1952-cc}, or the simpler Izhikevich one \citep{Izhikevich2003-rk}. However, most applications within the field of Deep Learning use simpler neuron models. Although less biologically realistic, they are of greater computational efficiency, which facilitates their application in large networks. Among those models, Leaky-Integrate-and-Fire (LIF) \citep{Dutta2017} is the most widely used. 

A LIF neuron consists of a leaky resistor in a parallel combination with a capacitor. Its dynamics can be described for a single neuron with the differential equation presented in Eq. \ref{lif}:

\begin{equation}
C \frac{d V}{d t}= -g_L(V(t)-E_L)+I(t) 
\label{lif}
\end{equation}

\noindent where $C$ is a constant representing the neuron capacitance; $V$ is the membrane potential; $g_L$ corresponds to another constant representing conductance, $E_L$ is the resting potential and $I(t)$ represents the input current. Specifically, both C and $g_L$ are decay parameters.

When the membrane potential, $V$, reaches or surpasses a pre-fixed threshold, a spike is generated, and $V$ changes to the neuron resting potential.

\subsubsection{Training Methods}
\label{stdp_related}

The discontinuous nature of SNN outputs prevents the use of backpropagation for SNN training. Hence, training SNNs has for some time been a challenge, and a number of alternative training methods have been developed to address it. Those training methods can be classified into the following three types \citep{Lan2022PCSNNSL}: conversion methods, that rely on the transformation of an already trained traditional ANN into an equivalent SNN; adapted backpropagation, where different techniques are applied to approximate backpropagation on SNNs despite the discrete nature of spikes, and local learning, bio-inspired methods where updates on connection weights are performed by using only locally accessible information to neurons.

One of the most well-known local learning methods is the standard Spike-Time Dependent Plasticity (STDP) \citep{Legenstein2008ALT}. This is an unsupervised learning rule in which the relative timing of spikes determines how synaptic connections are modified. Specifically, if we consider a connection between two neurons, the way in which that its synaptic weight is updated depends on the neurons firing order. If the presynaptic neuron fires shortly before the postsynaptic one, the connection between them is strengthened; whereas if the postsynaptic neuron fires before the presynaptic one, it is weakened. Concretely, if we consider the connection between two neurons, $X$, the presynaptic, and $Y$, the postsynaptic, the change in the strength of that connection, $\Delta \omega_{XY}$, is computed according to Eq. \ref{stdp}:
\begin{equation}
\label{stdp}
\Delta \omega_{XY} = 
\begin{cases}
A_+ \exp(-\Delta t / \tau_+), if \Delta t \geq 0 \\
A_- \exp(\Delta t / \tau_-), if \Delta t < 0
\end{cases}
\end{equation}
In Eq. \ref{stdp}, $\Delta t$ represents the time difference between the generation of a spike in $X$ and the generation of a spike in $Y$. That value will be positive if $Y$ spikes after $X$, and negative otherwise. $A_+$ and $A_-$ are parameters that regulate the strength of weight modifications, being $A_+$ usually positive, and $A_-$, usually negative; and $\tau_+$ and $\tau_-$ are constants that define the learning window for both reinforcement or weakening cases, respectively. 

\section{Proposed method}
\label{metodo}

In this paper, we introduce the Vacuum Spiker algorithm, an SNN approach for anomaly detection in univariate time series. Each input value is encoded as a single spike using Interval Coding and processed in real time by an SNN architecture. Trained exclusively on normal data using a modification of the STDP rule, the model learns to reduce its response to known patterns. Anomalies are detected when spike activity in the processing layer exceeds a threshold, offering an energy-efficient solution. 

\subsection{Interval Coding}
\label{codificacion}

To encode the input data, we propose a new encoding approach called Interval Coding. 
Let $D \subset \mathbb{R}$ be the initial domain of the input time series. This domain may correspond to a training set, a previously observed subset of the time series, or be defined based on prior knowledge. It is first partitioned into $k$ fixed-length, non-overlapping intervals, with each interval assigned to a unique neuron in the input layer. Together, these intervals cover the entire domain $D$. When a value $v$ is received at time $t$, the neuron corresponding to the interval that contains $v$ emits a spike. This mechanism slightly resembles population coding \citep{pan_neural_2019}; however, instead of employing a group of neurons with varying activation levels ---often in combination with other coding schemes--- a single spike from a single neuron is used to represent each input.

If an input value $v$ falls outside the current domain $D$, the domain is extended by appending contiguous, fixed-length, non-overlapping intervals to the boundary on the side where $v$ lies ---either below $\min(D)$ or above $\max(D)$---, until obtaining the smallest extended domain $D^* \supset D$ such that $v \in D^*$. Each of these additional intervals is assigned to a unique new input neuron, and together with the original ones they form a complete partition of $D^*$. To avoid creating an excessively large input layer, the extended domains $D^*$ can be bounded within a compact interval $I \subset \mathbb{R}$. In this case, if $v \notin I$, it is clamped to the nearest boundary of $I$.

This clamping step is crucial; otherwise, extreme values outside $I$ would not activate any neurons in the input layer, potentially leading to a drop in network activity and disrupting its functionality. Moreover, without clamping, such values could be mistaken for missing or no data. In Algorithm \ref{encoding}, the procedure for performing Interval Coding is presented step by step.

The motivation to choose the Interval Coding scheme is based on the limitations that STDP method can exhibit when used in combination with rate coding without complementary mechanisms. It has been shown that, under such conditions, STDP can approximate PCA \citep{2012_Gilson}, which is inherently a linear transformation. Thus, the proposed coding scheme is a strategy to prevent the SNN from falling in such a possible linear behaviour, which could limit its ability to capture complex patterns in data. The key idea is that, if the model generated by an SNN with rate coding could be approximated by a linear function depending on input data, an SNN with the proposed coding algorithm would be similar to a segmented linear regression, where a linear regression would approximate the SNN model for each interval. In this way, the model gains the capacity to approximate non-linear and intricate patterns, enhancing its potential to detect complex anomalies in the data.

It is also noteworthy that the proposed coding algorithm enables the coding of each single sample in just one time step, which removes the need for higher exposure times, that are common when other classical coding schemes are used. Coding each sample into a single time step not only facilitates the application of Vacuum Spiker algorithm in real time, but also enhances the applicability of the method to online learning scenarios, which could be further supported by the use of dynamically adapting domains $D^*$. Moreover, by relying on a single spike per sample, the approach significantly contributes to reducing the overall energy consumption of the model, which is particularly advantageous for resource-constrained systems.

\SetNoFillComment

\begin{algorithm}[H]
\caption{Interval Coding algorithm.}
\label{encoding}
\textbf{Input:} Univariate input series $V = input\_data$; number of intervals $k$; initial domain $D$; compact interval $I=[I_{min},I_{max}] \supset D$ \\
\textbf{Output:} Spike patterns to feed the SNN \\
Calculate length of intervals $\Delta=\max(V)-\min(V)$\;
$intervals \gets \text{Partition } D \text{ into } k \text{ non overlapping intervals of length } \Delta$\;
$neurons \gets \text{Dictionary with keys =} intervals$\;

\ForEach{$v \in V$}{
\If{$v < I_{min}$}{
$v \gets I_{min}$\;
}
\If{$v > I_{max}$}{
$v \gets I_{max}$\;
}

\While{$v \notin D$}{
\If{$v < \min(D)$}{
Append new interval $I_{new} = [\min(D)-\Delta, \min(D))$ to the left of $D$\;
Update $D \gets [\min(D)-\Delta, \max(D)]$\;
}
\If{$v > \max(D)$}{
Append new interval $I_{new} = [\max(D), \max(D)+\Delta)$ to the right of $D$\;
Update $D \gets [\min(D), \max(D)+\Delta]$\;
}
Add $I_{new}$ to $intervals$\;
Assign a unique new input neuron $n_{new}$ to $I_{new}$\;
Set $neurons[I_{new}]=n_{new}$\;
}

\ForEach{$b \in intervals$}{
\If{$v \in b$}{
$neurons[b]$ emits spike\;
\textbf{break}\;
}
}
}
\end{algorithm}

\subsection{Regulation of Synaptic Potentiation and Depression through STDP}
\label{general}

Building on the standard STDP formulation introduced in Section \ref{stdp_related}, this section explains how the dynamics of that learning method can be leveraged to control global synaptic behaviour.

Let two neurons be connected, $X$, the pre-synaptic one, and $Y$, the post-synaptic. According to standard STDP, if neuron $X$ fires before neuron $Y$, the connection between them becomes stronger. But, if $Y$ fires before $X$, the connection between them becomes weaker \citep{Legenstein2008ALT}. The amplitude of the modification of the weight corresponding to the connection between the two neurons, $\Delta \omega_{XY}$, is exponentially dependent on the time difference between the two firing events. It is also scaled by either one of two different constants, $A_+$ or $A_-$, depending on which neuron has fired first. 

Note that, in the STDP formulation, that can be seen in Eq. \ref{stdp}, the sign of the weight update $\Delta \omega_{XY}$ depends solely on the sign of $A_+$ and $A_-$. In standard STDP, $A_-$ is negative and $A_+$ is positive. However, in our approach both parameters are allowed to take either positive or negative values. In this way, depression events can be forced to dominate over potentiation events, or vice versa, regardless of the timing between pre- and post-synaptic spikes. Consequently, a connection can be shaped to either suppress or enhance its response to particular input patterns observed during training. This flexibility enables regulation of the global balance between synaptic potentiation and depression across the network, through an appropriate choice of $A_-$ and $A_+$ for the connections in it. By tuning these parameters, the Vacuum Spiker algorithm can be configured to display predominantly inhibitory dynamics in response to typical input patterns, thereby reducing network activity under normal conditions. Conversely, anomalous inputs may elicit a different response profile, characterized by increased activity. This dynamic provides the mechanism by which the Vacuum Spiker algorithm can modulate its activity based on the normality of the input data, thereby supporting its application to anomaly detection tasks.

\subsection{Vacuum Spiker Algorithm}
\label{vacuum_arq}

The proposed Vacuum Spiker algorithm combines two layers: $I$, the input layer, and $R$, the processing layer. A dense forward connection $I \rightarrow R$ is established. A dense recurrent connection $R \rightarrow R$ may also be present. In layer $I$, values from a monitored univariate time series are encoded using the Interval Coding scheme, described in \ref{codificacion}, and sent forward to $R$ as they are received. Layer $R$, composed of $n$ LIF neurons, can retain memory of past events through the membrane voltages of its neurons. 

The inference process for a single sample, detailed in Algorithm \ref{alg:inference}, begins by encoding the incoming value and propagating the resulting spikes to $R$. If the recurrent connection is present, previously generated spikes In $R$ are also fed back to $R$. Finally, the spiking activity in $R$ is recorded: if the number of firing neurons exceeds a threshold $\theta$, an alert is triggered; otherwise, no alert is raised.

Using the modified STDP rule described in \ref{general}, the model's connections are configured to exhibit a global inhibitory behaviour. This allows the Vacuum Spiker algorithm to learn to suppress or reduce its response to data similar to that found during training. Accordingly, the model is trained exclusively on normal data. When anomalies occur, the network exhibits increased spiking activity in layer $R$. If the number of firing neurons in $R$ surpasses a predefined threshold, an alert is triggered. Given that data is predominantly normal, this configuration may help to reduce the model's overall energy consumption.

The spiking nature of the model allows it to capture temporal dependencies directly through the dynamic evolution of neuron voltages in layer $R$, removing the need for sliding windows. This is because SNNs inherently process temporal information via discrete spike events, enabling each neuron's state to evolve in response to incoming stimuli. As a result, temporal patterns are encoded within the internal dynamics of the network itself, eliminating the need to explicitly segment input data into overlapping time windows, as is common in conventional neural networks. Additionally, the Interval Coding scheme avoids the need for exposure times longer than a single time step, which are commonly used in other SNN architectures. These properties significantly reduce the amount of data pre-processing required, compared with other time series algorithms, thereby facilitating the model’s applicability to real-world scenarios.

Together, the globally inhibitory configuration of the model and the efficiency of the Interval Coding scheme, contribute to a lower computational cost and improved energy efficiency of the Vacuum Spiker algorithm, making it a resource-conscious alternative for time series processing.

\begin{figure}[H]
 \centering
\includegraphics[width=\linewidth]{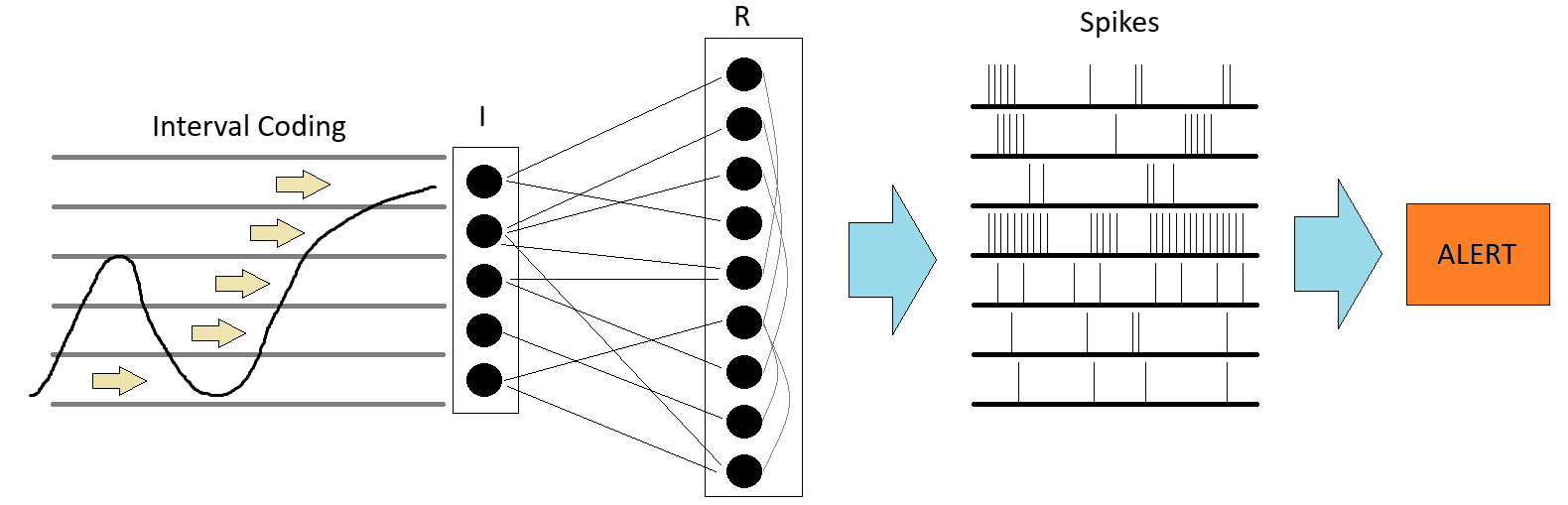}
\caption{Vacuum Spiker algorithm process. First, an univariate time series is encoded using Interval Coding. The predefined interval in which a value falls determines which neuron is activated. Information then propagates through the $I \rightarrow R$ connection (black lines between $I$ and $R$) and, if present, through $R \rightarrow R$ connection (gray lines within layer $R$). If the number of firing neurons in layer $R$ exceeds a predefined threshold, an alert is generated.}
\label{vacuum_proceso}
\end{figure}

In Fig. \ref{vacuum_proceso}, the overall anomaly detection process using the Vacuum Spiker algorithm can be found.

\begin{algorithm}[H]
\caption{Single time step inference for the Vacuum Spiker algorithm.}
\label{alg:inference}
\textbf{Input:} New value $v$ of a time series $V$\\
\textbf{Output:} $0$ if there is no alert; $1$ other case.

spikes from $I \gets \text{Interval Coding}(v)$\;
Propagate spikes from $I$ to $R$\;
\If {$R \rightarrow R$ exists}{
Propagate spikes from $R$ to $R$\;
}
Record $S \gets \text{spikes generated in } R$\;
\If{$\sum S>\theta$}{
\Return $1$\;
}\Else{
\Return $0$\;
}
\end{algorithm}

\section{Experimental setup}
\label{experimental}

The Vacuum Spiker algorithm has been trained and tested on a large collection of datasets, which are listed in Subsection \ref{datasets}. The same procedure was applied to the baseline algorithms described in Subsection \ref{baselines}. In Subsection \ref{energia}, the calculations performed to estimate the energy consumption for each model are described. Details about the execution of the experimentation are discussed in Subsection \ref{procedimiento}, and the metrics to evaluate the algorithms performance, and to choose the best configuration for the trained models, are elaborated in Subsection \ref{metricas}.

\subsection{Baseline algorithms}
\label{baselines}

As baseline algorithms, several anomaly detection models, obtained from the literature, were used. Also, some widely employed approaches are evaluated:

\begin{itemize}

\item \textit{\textbf{Convolutional Autoencoder (CAE)}} \citep{YILDIRIM2019121}: Model that combines the autoencoder architecture with one-dimensional convolutional layers. It has been developed to get a high performance, upper than $99.0\%$, while computational cost is kept low, through a technique to perform time series compression.

\item \textit{\textbf{Convolutional LSTM (CNN-LSTM)}} \citep{OH2018278}: Combination of convolutional layers with an LSTM. It has achieved a $98.5\%$ of accuracy in anomaly classification in electrocardiogram data.

\item \textit{\textbf{Deep Neural Network (DDNN)}} \citep{CAI2020103378}: Deep architecture that combines several blocks that compress and expand data. It has been applied to detect atrial fibrillation, with very high accuracy, sensitivity and specificity ($99.35 \pm 0.26\%$, $99.19 \pm 0.31\%$ and $99.44 \pm 0.17\%$, respectively). But, DDNN is also the most computationally expensive algorithm used in this study.

\item \textit{\textbf{1d-CAE-DL}} \citep{9925612}: Autoencoder architecture based on several convolutional layers. It outperforms other approaches in anomaly detection in batch manufacturing, achieving an Area Under the Curve (AUC) close to $100\%$.

\item \textit{\textbf{LSTM Autoencoder (LSTM-AE)}} \citep{GithinjiM23}: LSTM-based autoencoder. The reconstruction error is computed for each observation instead of each window. It outperforms traditional anomaly methods in anomaly detection in water level sensors data. It achieves an accuracy of $99\%$, along with a F1 score of $ 98.4\%$, a precision close to 1, and a recall of $96.9\%$ 

\item \textit{\textbf{One-Class SVM (OCSVM)}} \citep{s24082633}: Support Vector Machine technique that learns a decision boundary around normal data, considering anomalous the data that fall out such border. We used the Radial Basis Function Kernel (RBF), which is a popular choice in support vector machines.

\item \textit{\textbf{Local Outlier Factor (LOF)}} \citep{Auskalnis2018ApplicationOL}: Density-based anomaly detection algorithm that identifies anomalies by comparing the local density of a point with that of its neighbours. Instances that have significantly lower density than their neighbours are considered outliers. LOF is particularly robust to variations in data distribution.

\end{itemize}

\subsection{Datasets}
\label{datasets}

The following publicly available, well-known datasets have been used to evaluate the proposed model:

\begin{itemize}

\item \textit{\textbf{Dodgers Loop Sensor}} \citep{dodgers}: Loop sensor data collected from a traffic sensor close to the stadium of Dodgers in Los Angeles. Days when matches were celebrated in that stadium are accounted as anomalies to detect.

\item \textit{\textbf{CalIt2}} \citep{calit2}: People flow in and out the CalIt2 building, at the University of California, Irvine. Days when events were celebrated in that building are labelled as the anomalies to detect. Data has been separated in the two different cases contained in this dataset: people entering the building, and people leaving it.

\item \textit{\textbf{Numenta Anomaly Benchmark (NAB)}} \citep{AHMAD2017134}: A collection of 58 datasets covering different scenarios, most of them, coming from real world. These scenarios include different cases, such as such as CPU utilization, temperature measurements, Twitter volume, traffic speed, \textit{etc.} Out of the total, 52 datasets contain labelled anomalies.

\end{itemize}

The models were applied to each of the 52 datasets in the Numenta collection that contain labelled anomalies, as well as to the two different cases in CalIt2 ---people entering and exiting the building--- separately, in addition to the Dodgers dataset. In total, fifty five datasets were employed in the evaluation, encompassing a diverse range of time series scenarios. Every dataset has been preprocessed so that its records are separated by a constant time interval, while retaining the maximum possible number of records.

For nine datasets from the Numenta collection, at least one baseline model could not be applied because sufficiently long sequences of normal data for training were not available. For another dataset from the same collection, no model could be applied at all. Consequently, these ten datasets were excluded from the experiments, which were conducted on the remaining 45 datasets.

\subsection{Energy Consumption Estimation in Inference}
\label{energia}

To estimate the energy consumption of the traditional ANN models, the Vacuum Spiker algorithm, and the traditional machine learning algorithms used as baselines, we have followed the methodology exposed in \citep{9522999}. These estimations have been performed by counting the number of MAC operations required when inference was carried out on a single sample. Sums and products that could not be combined with others in a single MAC operation were treated as individual MAC operations. Non-linear activation functions such as sigmoid or hyperbolic tangent were not considered in this estimation. The energy required for each single operation performed on a device can be multiplied by the total number of MAC operations, to obtain the final estimation. Like that, the total number of MAC operations is roughly proportional to the energy consumption, and it can be used to compare the energy efficiency of different models.

The computations used to estimate the number of MAC operations required by each layer of the deep learning baselines, as well as by the traditional machine learning models, are presented in \ref{apendice}.

\subsubsection{Vacuum Spiker algorithm}

As established in \citep{9522999}, MAC operations in an SNN are of two types: 
\begin{itemize}
\item \textit{$E_s$}: Operations due to spikes crossing the neural connections.
\item \textit{$E_u$}: Operations due to the voltage updates performed in neurons along time.
\end{itemize}

As the implementation of the Vacuum Spiker algorithm is discrete, and it works in real time without any windows, one operation of kind $E_u$ is performed each time step for each neuron in the layer R. Therefore, the number of operations of kind $E_u$, i. e., voltage updates, performed each time step can be expressed as in Eq. \ref{eu}:
\begin{equation}
E_u = n
\label{eu}
\end{equation}
\noindent where n is the number of neurons in the layer $R$.

With respect to the operations required when a spike is produced, there are two kinds of spikes in Vacuum Spiker algorithm:

\begin{itemize}
\item Spikes arising from input layer $I$ and arriving to layer $R$.
\item Spikes produced in $R$.
\end{itemize}

By the way input data coding works in the Vacuum Spiker algorithm, only one spike is generated to code each new value arriving to the model. As every neuron in the layer $I$ is connected to every neuron in $R$, the number of operations due to those spikes equals the number of neurons in $R$, each time step. 

Regarding the spikes generated in $R$, if the model does not incorporate a recurrent connection, no MAC operations are necessary, as these spikes are not propagated to subsequent layers. But, if such connection exists, each spike produced in $R$ will require also one MAC operation for each neuron in $R$, due to the dense nature of the connection. Therefore, the number of MAC operations of kind $E_s$ performed for each time step can be written as in Eq. \ref{es}:

\begin{equation}
E_s=
\begin{cases}
n + n s_r & \text{if recurrence} \\
n & \text{if not recurrence}
\end{cases}
\label{es}
\end{equation}

\noindent where $n$ is the number of neurons in layer R, and $s_r$, the number of spikes generated in layer $R$. 

Finally, we get the total number of MAC operations performed by time step by adding $E_u$ and $E_s$. The final used expressions can be seen in Eq. \ref{vacuum_macs}:
\begin{equation}
M_V = E_u + E_s =
\begin{cases}
n(s_r + 2) & \text{if recurrence} \\
2n & \text{if not recurrence}
\end{cases}
\label{vacuum_macs}
\end{equation}

\subsection{Measuring the performance of algorithms}
\label{metricas}

There are different metrics used in literature to evaluate anomaly detection methods in time series. The scarcity of anomalies in data makes recommendable to take into account not only how well an algorithm can detect an anomaly, but its ability not to generate alarms when there are not anomalies. To address this issue, several approaches have been proposed \citep{KHORSHIDI20211}:
\begin{itemize}
\item \textit{Area Under the Curve (AUC)\citep{9247440}}: Area under the Receiver Operating Characteristic (ROC) curve, which represents the true positive rate (TPR) against the false positive rate (FPR), for each possible threshold. These two rates are defined as in Eqs. \ref{tpr_eq} and \ref{fpr_eq}:
\begin{itemize}
\item \begin{equation}
TPR=\frac{TP}{TP+FN}
\label{tpr_eq}
\end{equation}
\item \begin{equation}
FPR=\frac{FP}{TN+FP}
\label{fpr_eq}
\end{equation}
\end{itemize}
\noindent where $TP$ is the number of true positives, $FN$, the number of false negatives, $TN$, the number of true negatives and $FP$ the number of false positives. Specifically, the AUC is obtained by integrating $TPR$ as a function of $FPR$ over all possible thresholds:
\begin{equation}
AUC=\int_0^1 TPR d FPR
\end{equation}
An AUC of $0.5$ corresponds to a bad adjustment, while an AUC closer to 1 indicates a good adjustment between the predictor and the real class. This metric does not require the selection of a threshold to evaluate a model.
\item \textit{G-Mean\citep{2017-rao-mri11}:} Square root of the product of the $TPR$ by the true negative rate ($TNR$), where $TNR$ is calculated as in Eq. \ref{tnr_eq}:
\begin{equation}
FPR=\frac{TN}{TN+FP}
\label{tnr_eq}
\end{equation}
Like that, the G-Mean is calculated as exposed in Eq. \ref{G-Mean}:
\begin{equation}
\text{G-Mean}=\sqrt{TPR \cdot TNR}
\label{G-Mean}
\end{equation}
A G-Mean close to 0 indicates that the algorithm is unable to detect either negative or positive cases, falling into a trivial behaviour. A perfect match between algorithm and reality would correspond to a G-Mean of 1. 

\item \textit{F1-score\citep{s23031310}}: It is the ratio between the product of the obtained precision by the recall, and the sum of them. It is calculated by following the Eq. \ref{f1}:
\begin{equation}
F1=2\frac{P \cdot TPR}{P+TPR}
\label{f1}
\end{equation}
\noindent where $P$ is the precision, defined as in Eq. \ref{precision}:
\begin{equation}
P=\frac{TP}{TP+FP}
\label{precision}
\end{equation}

\end{itemize}

\subsection{Training and Evaluation Procedure}
\label{procedimiento}

For each dataset, expanding-window time series 5-fold cross-validation \citep{Kumar2017Time} was applied. For traditional machine learning and deep learning models, input data was standardized using z-score normalization. Anomalies were excluded from training datasets. For the deep learning baseline algorithms, the reconstruction error ---defined as the mean squared error between the true input and the model's reconstructed input--- has been used as the metric to determine whether a value is anomalous. For traditional machine learning algorithms, the predicted class labels generated by the models were used instead. For the Vacuum Spiker algorithm, the number of spikes generated in the layer $R$ along time has been the metric to decide the same. In that way, if either the reconstruction error, or the number of spikes in $R$, surpass a threshold ---set during hyperparameter tuning to maximize performance on a validation set--- it is considered that an anomaly could be happening. The experimentation was performed using Pytorch \citep{10.5555/3454287.3455008} , for traditional ANN models, and the library specialized on SNNs bindsnet \citep{10.3389/fninf.2018.00089}, for the Vacuum Spiker. Following the training procedure exposed in Subsection \ref{general}, constants $A_-$ and $A_+$ were set to different combinations of positive and negative values, for both the connections $I \rightarrow R$, and $R \rightarrow R$, covering a large set of potentiation and depression learning behaviours. The recurrent connection was optionally omitted. A grid search was performed using the parameter ranges specified in Table \ref{parametros}. The initial domain $D$ was set to the training set domain for each time series, and the compact interval bounding the subsequent domains $D^*$ was defined as $I = [2\min(D)-\max(D),2\max(D)-\min(D)]$. In this manner, the Vacuum Spiker algorithm was trained exclusively on normal data, with the objective of suppressing spiking activity in response to familiar input patterns rather than minimizing a global loss function. Within each dataset, the interval between consecutive records was kept constant. The exposure time for each record was set to 1 millisecond, which corresponded to the simulation time step.

The parameters of the baseline models are also presented in Table \ref{parametros}. Window size was used for models without a predefined value for this parameter, i.e., LSTM-AE, 1d-CAE-DL, OCSVM, and LOF. Hidden layers sizes and latent dimensions were employed for LSTM-AE. Additionally, the number of layers was considered only for the LSTM-AE and 1-CAE-DL models.

Before evaluating the performance, the anomaly detection signals ---whether reconstruction errors, spike counts or predicted class labels--- were optionally smoothed using a moving average over the previous 100, 200, or 300 records. The best result obtained for across all smoothing windows, including the case without smoothing, was retained. The motivation behind this smoothing step is to avoid unfair penalization of the Vacuum Spiker algorithm. In this model, anomalies may manifest as an increased number of spikes distributed over time, however, this doesn't necessarily imply elevated activity in every individual record within an anomalous segment, where alternating patterns of spiking and non-spiking activity are often observed. As a result, without smoothing, the discrete and temporally sparse nature of the spikes could lead to underestimation of the algorithm's ability to detect anomalous patterns.

To evaluate performance, the three metrics described in Section \ref{metricas} were applied. For G-Mean and F1-score, various thresholds were tested on each anomaly detection signal to determine whether each point should be classified as an anomaly. For each signal, the threshold yielding the best performance was selected. Ten threshold values were used, uniformly spaced between the minimum and maximum values of each anomaly detection signal. For each evaluation metric (G-Mean, AUC, and F1-score), the best-performing configuration was selected independently. Therefore, the reported results for each metric correspond to the optimal model identified for that specific evaluation criterion. In cases where no true positives were detected or no positive predictions were made, the F1-score was assigned a value of zero.

\begin{table}[h]
 \footnotesize
 \centering
 \caption{Parameters used in the grid search.}
 \label{parametros}
 \begin{tabular}{|c|c|c|}
\toprule
\textbf{Kind of model} & \textbf{Parameter} & \textbf{Values}\\ 
\midrule
\multirow{14}{*}{SNN}
& $A_-$ for $I \rightarrow R$ & $[-0.1,0.1]$\\
& $A_+$ for $I \rightarrow R$ & $[-0.1,0.1]$\\
& $A_-$ for $R \rightarrow R$ & $[-0.1,0.1]$ \\
& $A_+$ for $R \rightarrow R$ & $[-0.1,0.1]$\\
& Weight initialization for $I \rightarrow R$ & $\mathcal{N}(0.05,0.1)$\\
& Weight initializacion for $R \rightarrow R$ & $0.025 \cdot (\mathbb{I}-1)$\\
& Num. of neurons in $R$ & $[100,2000]$\\
& Spike threshold & $[-62,-55,-40]$ mV\\
& $g_L$ & $[1-e^{-1/100},1-e^{-1/150},1-e^{-1/200}]$\\
& Interval size & $[0.1, 10]$ $\%$ of training domain\\
& Neurons resting potential & $-65$ mV\\ 
& Neurons reset potential & $-65$ mV\\ 
& Neurons refractory period & $5$ ms\\
& STDP $\tau_+$ and $\tau_-$ & $1.051$ ms\\
& Neurons' $C$ constant & $1$ $\mu F$\\
& Epochs & $[1,2,3,4,5]$\\
\midrule
\multirow{8}{*}{Baseline}
& Batch size & $[32,64,128]$\\
& Learning rate & $[0.001,0.005,0.01,0.1]$\\
& Window size & $[10,50,100,150,200]$\\
& Sizes of the hidden layers & $[32,64]$\\
& Latent dimensions & $[50,100]$\\
& Num. of layers & $[1,2,3]$\\
& $\nu$ (OCSVM) & $[0.05,0.2]$\\
& Num. of neighbours (LOF) & $[30,50]$\\
& Epochs & $[10,50,100]$\\
\bottomrule
 \end{tabular}
\end{table}

The number of MAC operations has been used as an estimator of the energy consumption of the various evaluated approaches. For traditional ANN models, the number of required MAC operations was calculated across the layers of each model, following the formulas presented in Section \ref{energia}. To estimate the energy consumption of the Vacuum Spiker algorithm and the traditional machine learning algorithms, the methodology described in the same section was applied. In every case, the reported MAC count corresponds to the specific model configuration that achieved the best performance for the respective evaluation metric (G-Mean, AUC, or F1-score). In the case of a tie in performance, the configuration requiring the fewest MAC operations was selected.

This evaluation pipeline was applied to the datasets mentioned in \ref{datasets}. The Iman-Davenport test \citep{Iman01011980} was employed to assess whether there are statistically significant differences among the tested algorithms in both performance and energy consumption. When such significant differences were identified, the Wilcoxon signed-rank test \citep{Wilcoxon1992}, accompanied by Holm's correction \citep{29def780-e117-38f0-8afb-edf384af3fad}, was subsequently applied to determine which specific algorithms exhibited significant differences in performance and energy usage with respect to the Vacuum Spiker.

Additionally, the various tested parameter combinations $(A_-, A_+)$ for both the $I \rightarrow R$ and $R \rightarrow R$ connections were classified according to the predominant effect they trend to induce on synaptic weights during training ---namely, excitatory (potentiation-dominated), inhibitory (depression-dominated), or balanced (where potentiation and depression are approximately equal). For each dataset, only the parameter combination that achieved the best performance was considered. These selected combinations were subsequently analysed using a chi-squared ($\chi^2$) test to evaluate whether excitatory, inhibitory, and balanced configurations occurred with equal frequency across all the datasets, or whether certain types appeared significantly more often, suggesting the presence of a dominant configuration that could be better suited for the anomaly detection task with the Vacuum Spiker algorithm.

Since the STDP parameters $\tau_+$ and $\tau_-$ were set to the same value, and it was sufficiently large to allow substantial weight updates for spikes separated by relatively long time intervals, the prevalent behaviour induced in connections during training has been estimated by examining the sign and magnitude of the parameters $A_+$ and $A_-$, and the number of spikes generated in pre- and post-synaptic layers, by following the reasoning outlined below.

For a dense connection $L_1 \rightarrow L_2$, each time a neuron $X \in L_1$ spikes, weight updates are applied to those connections from $X$ to neurons in $L_2$ that spiked earlier. These updates are scaled by $A_-$. Similarly, each time a neuron $Y \in L_2$ spikes, weight updates are applied to connections from neurons in $L_1$ to $Y$, scaled by $A_+$. Consequently, if $A_-=-A_+$, the layer with the higher firing activity tends to determine the prevalent behaviour of the connection: if $L_1$ spikes more frequently, the sign of $A_-$ would dominate (depression if negative, potentiation if positive), whereas if $L_2$ spikes more, the sign of $A_+$ would dominate.

Therefore, if both $A_-$ and $A_+$ are both positive (negative), potentiation (depression) is globally favoured in the connection $L_1 \rightarrow L_2$ during learning. If $A_-=-A_+$ and $L_1 \rightarrow L_2$ is recurrent, with $L_1=L_2$, the number of spikes generated in both layers over time is equal, and the behaviour of the network would be approximately balanced. If $L_1 \rightarrow L_2$ is a dense forward connection, the layer that generates more spikes would determine the tendency of the connection's prevalent behaviour. In the case of the Vacuum Spiker algorithm, the balance between low weight initialization and the employed spike thresholds leads layer $R$ to tend to generate fewer spikes than $I$. For this reason, in the case that $A_-=-A_+$, the sign of $A_- $ would roughly define of the $I \rightarrow R$ prevalent behaviour.

\section{Results and Discussion}
\label{resultados}

The Iman-Davenport test yielded a p-value of $0$ for performance across all metrics used ---G-Mean, AUC, and F1-score. It also yielded a p-value of $0$ for energy consumption for the models selected according to these metrics. In Table \ref{iman-davenport}, the p-values obtained from the Iman-Davenport test can be seen. This result provides strong evidence supporting the hypothesis that there are significant differences in both performance and energy consumption among the evaluated models, thus justifying the application of the signed-rank Wilcoxon test with Holm correction to analyse these differences. 

\begin{table}
 \footnotesize
 \centering
 \caption{p-values from Iman-Davenport test for G-Mean, F1-score and AUC and energy consumption.}
 \label{iman-davenport}
 \begin{tabular}{|c|c|c|c|}
\toprule
\textbf{} & \textbf{G-Mean} & \textbf{F1-score} & \textbf{AUC}\\ 
\midrule
Performance & 0.0000 & 0.0000 & 0.0000\\
Number of MACs & 0.0000 & 0.0000 & 0.0000\\
\bottomrule
 \end{tabular}
\end{table}

Accordingly, such test was applied to the best values obtained for G-Mean, AUC, and F1-score on each individual dataset, as well as to the corresponding number of required MACs (Multiply-Accumulate Operations). The resulting p-values related to performance are presented in Table \ref{performance_p}, while the median performance values for each model are shown in Table \ref{performance_m}. P-values for energy consumption are included in Table \ref{energia_p}, and the median energy consumption values are reported in Table \ref{energia_m}.

As shown in Tables \ref{performance_p} and \ref{performance_m}, the Vacuum Spiker algorithm performed statistically significantly better than the machine learning models LOF and OCSVM, as well as the deep learning approaches CNN-LSTM and DDNN, across the three performance metrics employed. These baseline models exhibited the lowest performance across all metrics.

On the other hand, the Vacuum Spiker algorithm achieved the highest median G-Mean across all datasets, with no statistically significant differences compared to CAE, LSTM-AE, and 1d-CAE-DL. In terms of median F1-score, the proposed algorithm was ranked third, with the only significant difference observed relative to the top-performing model, LSTM-AE. Regarding AUC, Vacuum Spiker also was ranked third, but the differences with the two top-performing models were not statistically significant.

Overall, the obtained results indicate that the Vacuum Spiker algorithm consistently ranked among the top-performing models, demonstrating competitive performance across diverse metrics and datasets. These findings suggest that the algorithm can provide a robust and reliable alternative for anomaly detection on time series, performing comparably to several well-established methods under the evaluated conditions. 

Regarding energy efficiency, as shown in Table \ref{energia_m}, the Vacuum Spiker algorithm significantly outperforms all baseline models, except LOF, in terms of median MAC operations. However, LOF consistently yields the lowest performance across all evaluation metrics (G-Mean, F1-score, and AUC). The reduced computational cost of the Vacuum Spiker algorithm can be attributed to specific design choices, including the use of an efficient input coding scheme that emits only one spike per input value, and the absence of time windowing in the data processing pipeline. These elements reduce the number of spike-triggered computations per time step, thereby significantly reducing the number of required MAC operations. In this way, they contribute to a favourable trade-off between anomaly detection performance and energy consumption.

\begin{table}[H]
 \footnotesize
 \centering
 \caption{Adjusted p-values, with Holm correction, from Wilcoxon signed-rank tests comparing Vacuum Spiker with baseline models in terms of G-Mean, F1-score, and AUC. Bold values indicate statistical significance at $\alpha = 0.05$.}
 \label{performance_p}
 \begin{tabular}{|c|c|c|c|}
\toprule
\textbf{Comparison}& \textbf{p-value (G-Mean)} & \textbf{p-value (F1-score)} & \textbf{p-value (AUC)}\\
\midrule
Vacuum Spiker vs CAE & $5.3548 \cdot 10^{-2}$ & $1.9470 \cdot 10^{-1}$ & $5.7705 \cdot 10^{-1}$ \\
Vacuum Spiker vs CNN-LSTM & $\mathbf{4.3199 \cdot 10^{-8}}$ & $\mathbf{4.8193 \cdot 10^{-8}}$ & $\mathbf{3.1991 \cdot 10^{-6}}$ \\
Vacuum Spiker vs DDNN & $\mathbf{1.1372 \cdot 10^{-4}}$ & $\mathbf{1.6253 \cdot 10^{-3}}$ & $\mathbf{1.8787 \cdot 10^{-3}}$ \\
Vacuum Spiker vs 1d-CAE-DL & $1.0000 \cdot 10^{0}$ & $6.4659 \cdot 10^{-1}$ & $1.0000 \cdot 10^{0}$ \\
Vacuum Spiker vs LSTM-AE & $1.0000 \cdot 10^{0}$ & $\mathbf{1.9851 \cdot 10^{-2}}$ & $1.0000 \cdot 10^{0}$ \\
Vacuum Spiker vs LOF & $\mathbf{4.5293 \cdot 10^{-9}}$ & $\mathbf{8.2366 \cdot 10^{-11}}$ & $\mathbf{8.0973 \cdot 10^{-10}}$ \\
Vacuum Spiker vs OCSVM & $\mathbf{2.8549 \cdot 10^{-5}}$ & $\mathbf{7.3326 \cdot 10^{6}}$ & $\mathbf{2.6664 \cdot 10^{-5}}$ \\
\bottomrule
 \end{tabular}
\end{table}

\begin{table}[H]
 \footnotesize
 \centering
 \caption{Median performance values for each model in terms of G-Mean, F1-score, and AUC. Best results for each metric are shown in bold.}
 \label{performance_m}
 \begin{tabular}{|c|c|c|c|}
\toprule
\textbf{Model} & \textbf{G-Mean (median)} & \textbf{F1-score (median)} & \textbf{AUC (median)} \\
\midrule
Vacuum Spiker & \textbf{0.8645} & 0.7640 & 0.8649 \\
CAE & 0.7658 & 0.7137 & 0.8430 \\
CNN-LSTM & 0.6700 & 0.6519 & 0.7281 \\
DDNN & 0.7101 & 0.6949 & 0.7821 \\
1d-CAE-DL & 0.8377 & 0.7723 & \textbf{0.8937} \\
LSTM-AE & 0.8542 & \textbf{0.8362} & 0.8833 \\
LOF & 0.6403 & 0.5022 & 0.6373 \\
OCSVM & 0.7032 & 0.6401 & 0.6912 \\
\bottomrule
 \end{tabular}
\end{table}

\begin{table}[H]
 \footnotesize
 \centering
 \caption{Adjusted p-values with Holm correction, from Wilcoxon signed-rank tests comparing Vacuum Spiker with baseline models in terms of number of the number of MACs required to perform inference in a single sample with each model, selected based on best G-Mean, F1-score or AUC. Bold values indicate statistical significance at $\alpha = 0.05$.}
 \label{energia_p}
 \begin{tabular}{|c|c|c|c|}
\toprule
\textbf{Comparison}& \textbf{p-value (G-Mean)} & \textbf{p-value (F1-score)} & \textbf{p-value (AUC)}\\
\midrule
Vacuum Spiker vs CAE & $\mathbf{2.0688 \cdot 10^{-8}}$ & $\mathbf{2.0711 \cdot 10^{-8}}$ & $\mathbf{3.9790 \cdot 10^{-13}}$ \\
Vacuum Spiker vs CNN-LSTM & $\mathbf{1.0276 \cdot 10^{-7]}}$ & $\mathbf{7.0149 \cdot 10^{-8}}$ & $\mathbf{2.9655 \cdot 10^{-10}}$ \\
Vacuum Spiker vs DDNN & $\mathbf{2.0688 \cdot 10^{-8}}$ & $\mathbf{2.0711 \cdot 10^{-8}}$ & $\mathbf{3.9790 \cdot 10^{-13}}$ \\
Vacuum Spiker vs 1d-CAE-DL & $\mathbf{4.7748 \cdot 10^{-12}}$ & $\mathbf{9.9476 \cdot 10^{-12}}$ & $\mathbf{1.2187 \cdot 10^{-10}}$ \\
Vacuum Spiker vs LSTM-AE & $\mathbf{7.9581 \cdot 10^{-13}}$ & $\mathbf{9.4303 \cdot 10^{-10}}$ & $\mathbf{1.0544 \cdot 10^{-10}}$ \\
Vacuum Spiker vs LOF & $\mathbf{9.2760 \cdot 10^{-7}}$ & $\mathbf{5.7099 \cdot 10^{-6}}$ & $\mathbf{3.8338 \cdot 10^{-5}}$ \\
Vacuum Spiker vs OCSVM & $\mathbf{4.7910 \cdot 10^{9-}}$ & $\mathbf{2.4718 \cdot 10^{-9}}$ & $\mathbf{7.5056 \cdot 10^{-10}}$ \\
\bottomrule
 \end{tabular}
\end{table}

\begin{table}
 \footnotesize
 \centering
 \caption{Median number of MACs (in thousands) required to perform inference in a single sample with each model, selected based on best G-Mean, F1-score, or AUC. Lowest values per column are highlighted in bold.}
 \label{energia_m}
 \begin{tabular}{|c|c|c|c|}
\toprule
\textbf{Model} & \textbf{G-Mean selection} & \textbf{F1-score selection} & \textbf{AUC selection} \\
\midrule
Vacuum Spiker & 4.0544 & 4.1437 & 4.0595 \\
CAE & 992.7080 & 1543.5420 & 1543.5420 \\
CNN-LSTM & 31.6800 & 31.6800 & 31.6800 \\
DDNN & 1001.8480 & 1001.8480 & 1001.8480 \\
1d-CAE-DL & 413.4400 & 48.0000 & 48.0000 \\
LSTM-AE & 33.1520 & 33.1520 & 30.5920 \\
LOF & \textbf{1.1920} & \textbf{1.2920} & \textbf{1.2920} \\
OCSVM & 44.9208 & 74.6996 & 48.0420 \\
\bottomrule
 \end{tabular}
\end{table}

\begin{figure}[ht]
\centering
\begin{subfigure}[b]{0.45\linewidth}
 \centering
\includegraphics[width=\linewidth]{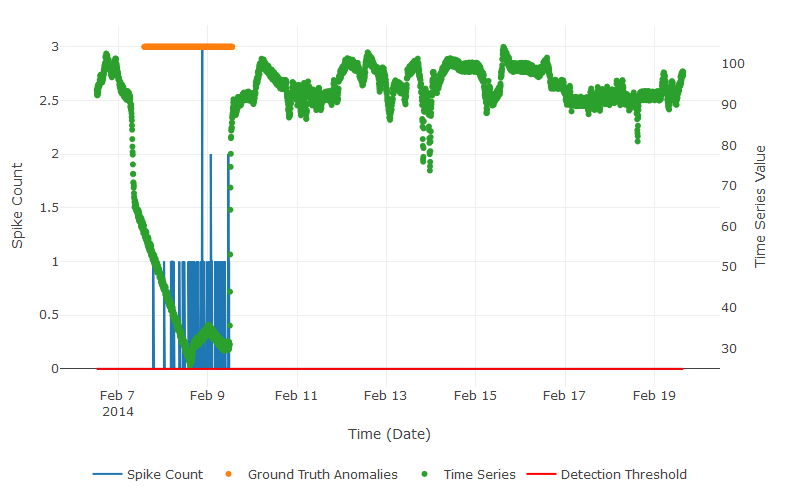}
\caption{\nolinkurl{realKnownCause\_machine\_temperature\_system\_failure}, from the Numenta collection. Best performing configuration according to the G-Mean metric.}
\label{figura_G-Mean}
\end{subfigure}
\hfill
\begin{subfigure}[b]{0.45\linewidth}
 \centering
\includegraphics[width=\linewidth]{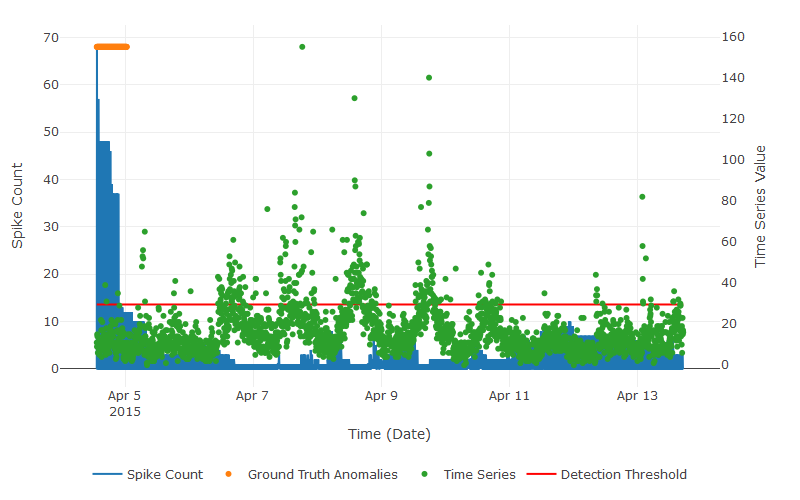}
\caption{\nolinkurl{realTweets\_Twitter\_volume\_FB}, from the Numenta collection. This configuration achieved the Best performing configuration according to the F1-score.}
\label{figura_f1}
\end{subfigure}
\hfill
\begin{subfigure}[b]{0.45\linewidth}
 \centering
\includegraphics[width=\linewidth]{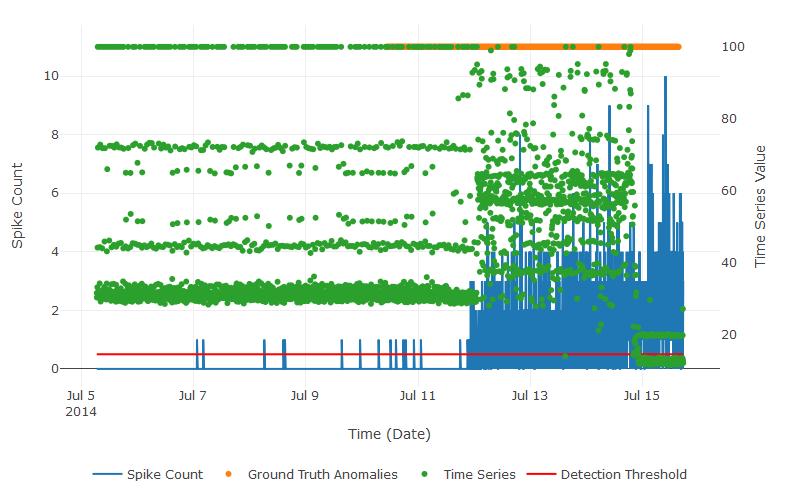}
\caption{\nolinkurl{realKnownCause\_cpu\_utilization\_asg\_misconfiguration}, from the Numenta collection. Best performing configuration according to the AUC metric. The anomaly detection threshold was estimated with the best Youden index.}
\label{figura_auc}
\end{subfigure}
\caption{Spike counts generated by the Vacuum Spiker algorithm (blue). Green dots indicate the time series values, while orange lines mark intervals with labelled anomalies. The red lines denote the anomaly detection thresholds.}
\label{figuras_resultados}
\end{figure}

In Figure \ref{figuras_resultados}, the response of the Vacuum Spiker algorithm is shown across several datasets used in our experiments. In these plots, green dots represent the values of the time series over time, orange horizontal lines indicate the time intervals during which ground truth anomalies are present. The blue lines correspond to the spike counts generated in layer $R$ of the Vacuum Spiker algorithm when processing the time series, and the red line represents the detection threshold, above which a potential anomaly is flagged. For the F1-score and G-Mean metrics, the threshold shown corresponds to the one that yielded the best performance on each respective dataset. In the case of AUC, the threshold was selected based on the maximum Youden Index \citep{hughes2015youden}.

\subsection{Analysis of Synaptic Behaviour}

To examine whether specific combinations of excitatory, inhibitory, and balanced synaptic behaviours in the $I \rightarrow R$ and $R \rightarrow R$ connections could be associated with superior performance more frequently than others, a statistical analysis of their occurrence as optimal configurations was conducted across datasets. To assess this, a $\chi^2$ goodness-of-fit test \citep{Balakrishnan2013ChiSquared} was applied to evaluate whether the frequency with which different combinations emerged as optimal for each performance metric deviates significantly from a uniform distribution.

Specifically, a separate $\chi^2$ test was conducted for each performance metric. Synaptic behaviours with frequencies lower than 5 were grouped together. The null hypothesis, $H_0$, stated that all combinations of synaptic behaviours had the same probability of occurrence. All the $54$ datasets that could be processed by the Vacuum Spiker algorithm were included in the analysis.

The tests yielded p-values of $0.012$ for G-Mean, $0.029$ for F1-score, and $2.806 \cdot 10^{-6}$ for AUC. These results suggest a statistically significant deviation from $H_0$, indicating that some combinations of synaptic behaviours occur with different frequencies. The p-values obtained from these tests are summarized in Table \ref{chi_p}.

\begin{table}
 \footnotesize
 \centering
 \caption{P-values from the $\chi^2$ goodness-of-fit test applied to each performance metric. Statistically significant values are shown in bold.}
 \label{chi_p}
 \begin{tabular}{|c|c|c|}
\toprule
\textbf{G-Mean} & \textbf{F1-score} & \textbf{AUC} \\
\midrule
\textbf{\boldmath$1.2023 \cdot 10^{-2}$} & \textbf{\boldmath$2.8726 \cdot 10^{-2}$} & \textbf{\boldmath$2.8063 \cdot 10^{-6}$} \\
\bottomrule
\end{tabular}
\end{table}

Table \ref{chi_tabla} presents the standardized residuals for each synaptic combination, according to the corresponding performance metric. The column Syn. $I \rightarrow R$ indicates the dominant synaptic behaviour in the feedforward connection, while Syn. $R \rightarrow R$ refers to the recurrent connection. Synaptic behaviours are abbreviated as Inh. (inhibitory), Exc. (excitatory), and Neu. (neutral). The remaining columns display the standardized residuals for each selection metric. Combinations with frequencies lower than 5 were grouped under ‘Other’, with the specific elements included varying by metric. Empty cells indicate synaptic combinations that were grouped under ‘Other’ for the corresponding selection metric.

As it can be observed in that table, the combination of an excitatory-prevalent synaptic behaviour in the forward connection $I \rightarrow R$, and an inhibitory-prevalent one in the recurrent connection $R \rightarrow R$, was the only configuration that appeared with a significantly higher frequency as the optimal across all the datasets. Its standardized residuals were $3.130$ for G-Mean and F1-score, and $5.511$ for AUC, all of them exceeding the threshold of $2$. This indicates that this configuration tends to occur more frequently than would be expected by chance. Specifically, it occurred $20$ times when G-Mean or F1-score was used as the performance metric, and $27$ times when AUC was considered, out of the $54$ datasets analysed, which suggest that this configuration could be particularly suitable for anomaly detection with the Vacuum Spiker algorithm.

A possible explanation for the predominance of the excitatory forward and inhibitory recurrent configuration ($I \rightarrow R$: Exc., $R \rightarrow R$: Inh.) lies in the type of predictive dynamics it induces within the network. When a value from the input time series is presented, the excitatory forward connection tends to activate a subset of neurons in layer $R$ that have become responsive not only to that specific input value but also ---through co-activation during training--- to other values that frequently co-occur with it in temporal proximity. As a result, inputs that tend to follow one another over time may converge onto overlapping subsets of neurons in $R$. This overlapping activation has functional implications under the influence of the recurrent inhibitory connection. Since active neurons inhibit each other via $R \rightarrow R$, the initial activation of co-responsive neurons leads to suppression of activity associated with likely subsequent inputs. Consequently, layer $R$ enters a transiently silent state, resuming activity only once inhibition decays.

In the case of an anomaly, the neurons responsive to it are likely to remain uninhibited, allowing them to fire and generate a new wave of inhibition adapted to the novel input. This results in a transient increase in network activity, reflecting a failure of the internal expectations encoded in the inhibitory dynamics. 

It could be considered that this configuration implements a form of suppressive prediction: anticipated future inputs are inhibited pre-emptively, while deviations from the expected pattern elicit enhanced responses. These dynamics are consistent with principles of predictive coding, where prediction errors drive changes in neural activity to update internal models in real time \citep{Millidge2021Predictive}.

Other configurations of synaptic behaviour may face limitations when applied to anomaly detection. For instance, excitatory-forward architectures lacking recurrent inhibition may produce homogeneous and persistent activation over time, as temporally co-occurring inputs are likely to activate overlapping subsets of neurons in layer $R$, resulting in reduced variability in responses. On the other hand, configurations with an inhibitory forward connection may suppress activation in layer $R$ so strongly that the training in the recurrent connection becomes irrelevant.

Interestingly, configuration with predominance of excitation in the forward connection and inhibition of the recurrent one, partially mirrors a well-known principle in the visual system, where lateral connections between spatially adjacent neurons are predominantly inhibitory \citep{Battaglini2019}. This anatomical motif is thought to support functions such as contrast enhancement and noise suppression, and may reflect a more general computational strategy for selectively amplifying unexpected or informative stimuli.

\begin{table}[H]
 \footnotesize
 \centering
 \caption{Residual analysis of synaptic behaviour combinations classified as optimal according to the G-Mean, F1-score, and AUC. Each column reports the standardized residuals obtained for the corresponding performance metric. Combinations with frequencies lower than 5 were grouped under ‘Other’. Standardized residuals significantly greater than expected (i.e., above 2) are highlighted in bold.}
 \label{chi_tabla}
 \begin{tabular}{|c|c|c|c|c|}
\toprule
\textbf{Syn. \boldmath$I \rightarrow R$} & \textbf{Syn. \boldmath$R \rightarrow R$} & \textbf{Res. (G-Mean)} & \textbf{Res. (F1-score)} & \textbf{Res. (AUC)} \\
\midrule
\textbf{Exc.} & \textbf{Inh.} & \textbf{3.1299} & \textbf{3.1299} & \textbf{5.5114}\\
Exc. & Neu. & -1.6330 & -1.2928 & -1.9732\\
Inh. & Exc. & -1.6330 & -1.2928 & -\\
Inh. & Neu. & -0.6124 & 0.0680 & -1.6330\\
Inh. & Inh. & - & - & -1.2928 \\
\hline
\multicolumn{2}{|c|}{Other} & 0.7485 & -0.6124& -0.6124\\
\bottomrule
\end{tabular}
\end{table}

\FloatBarrier

\section{Case Study: Malfunction of Photovoltaic Systems}
\label{solar}

In this section, we show how the Vacuum Spiker algorithm can be used to monitor the state of a unattended solar inverter at the edge.

The solar plant under consideration contains one inverter. To avoid a reduction in the energy output of the facility, it is crucial to detect any malfunction in that inverter as soon as possible, enabling timely repair or replacement. Since the facility is isolated and typically unmanned, there is a strong interest in implementing an anomaly detection system capable of autonomously responding to potential failures.

However, the facility owners are concerned that if the anomaly detection system relies on the solar plant itself for power, potential faults might go unnoticed, as the system could be shut down. This concern has led them to prefer that the anomaly detection system be powered by a battery. Nevertheless, relying on such a power source imposes strict limitations on the system’s energy efficiency. 

To address these constraints and maximize the system’s efficiency, the Vacuum Spiker algorithm was trained and tested for deployment in this context.

\subsection{Data}

To develop a model able to detect problems in the above-mentioned solar inverter, we have records taken each 5 minutes, corresponding to the power, in kW, generated by it over nine months. During the first five months, we know that the inverter operated correctly. However, during the following four months, it began to show power curtailment, which at first appeared sporadically, but gradually increased in frequency.

The available data consisted of two variables: a timestamp, composed by the date and time, and the average power production over the previous five minutes. The data was divided into training and test sets, with the first five months used for training and the following four months, during which power curtailment occurred, used for testing. Anomalies present in the data were labelled by a team of experts.

In total, there were $43687$ training records and $34479$ test records for the studied inverter.

Although the power output of a solar inverter typically follows a regular pattern, it can be influenced by meteorological conditions. Factors such as clouds, rain, and other environmental elements can introduce noise into the data, complicating the process of determining whether any performance reduction is due to an issue with the system or the surrounding conditions. 

To address this challenge, The Vacuum Spiker algorithm was trained using the data from the training set. The test set was then used to assess the models’ ability to detect deviations in power output under real operating conditions.

\subsubsection{Anomalies}

\begin{figure}[ht]
 \centering
 \begin{subfigure}[b]{0.45\linewidth}
\centering
\includegraphics[width=\linewidth]{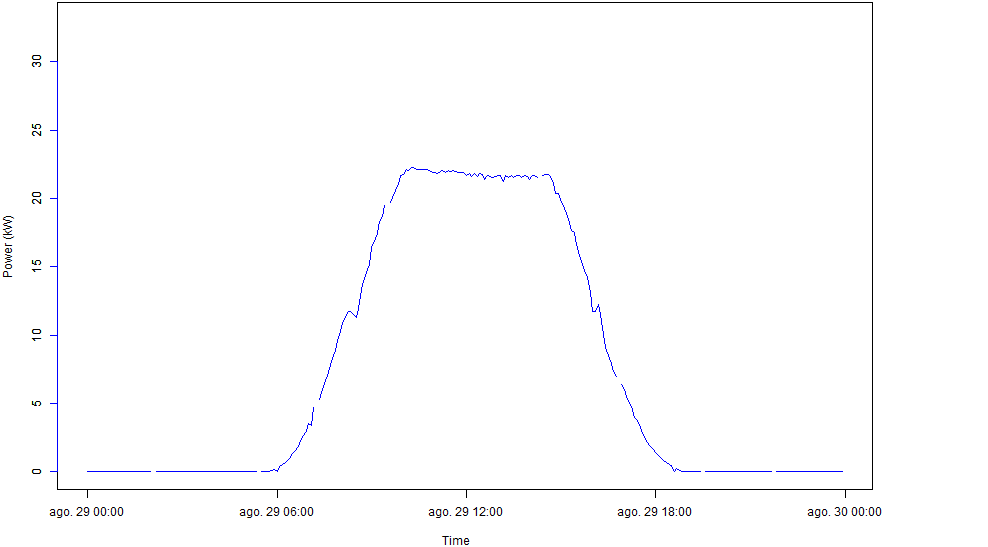}
\caption{Power limitation due to curtailment.}
\label{limitacion}
 \end{subfigure}
 \hfill
 \begin{subfigure}[b]{0.45\linewidth}
\centering
\includegraphics[width=\linewidth]{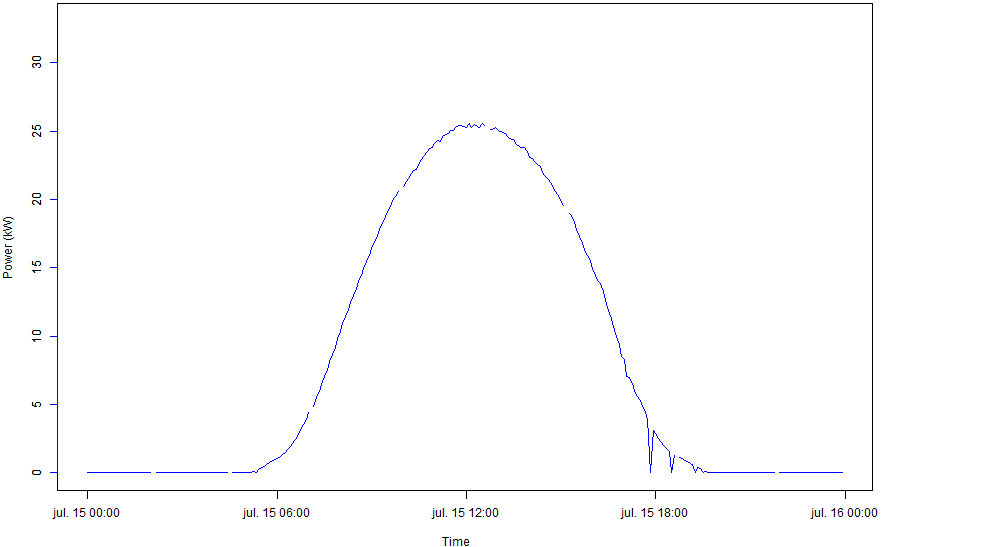}
\caption{Normal operation.}
\label{no_limitacion}
 \end{subfigure}
 \caption{Comparison of power output (kW) by the solar inverter with and without power curtailment.}
 \label{fig:produccion_comparada}
\end{figure}

The studied inverter has experienced power curtailment, meaning it was unable to reach its expected maximum production throughout the day. In Fig. \ref{limitacion}, the power output of the curtailed inverter is shown, over the course of a day. It can be observed that the inverter is unable to produce more than 23 kW, maintaining an almost constant output around 22 kW during the sunniest hours of the day. In contrast, Fig. \ref{no_limitacion} displays the behaviour of the same inverter on a day when no power curtailment occurred.

\begin{figure}[ht]
 \centering
 \begin{subfigure}[b]{0.45\linewidth}
\centering
\includegraphics[width=\linewidth]{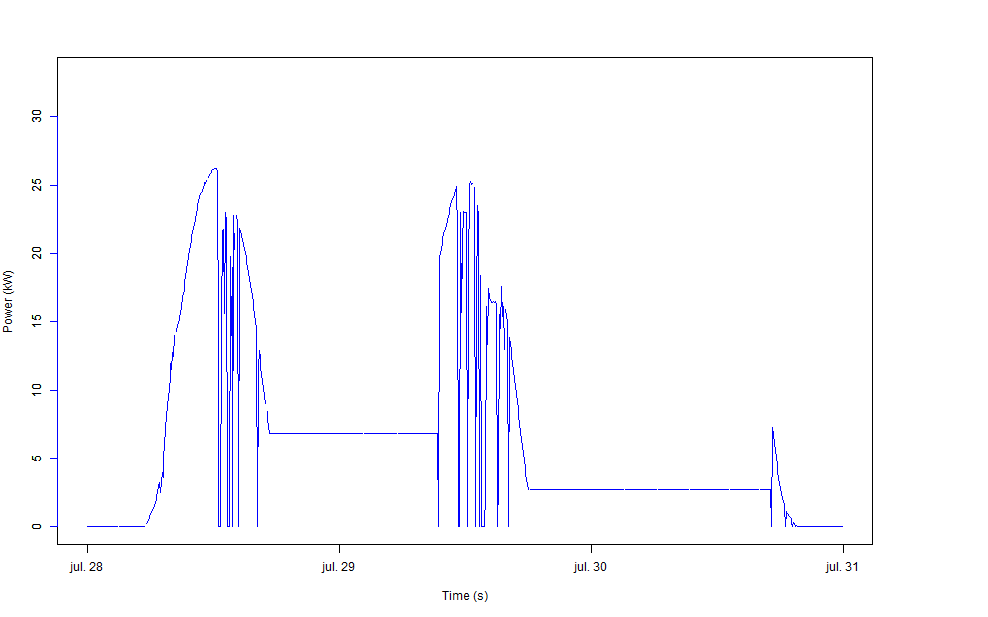}
\caption{Communication error.}
\label{error1}
 \end{subfigure}
 \hfill
 \begin{subfigure}[b]{0.45\linewidth}
\centering
\includegraphics[width=\linewidth]{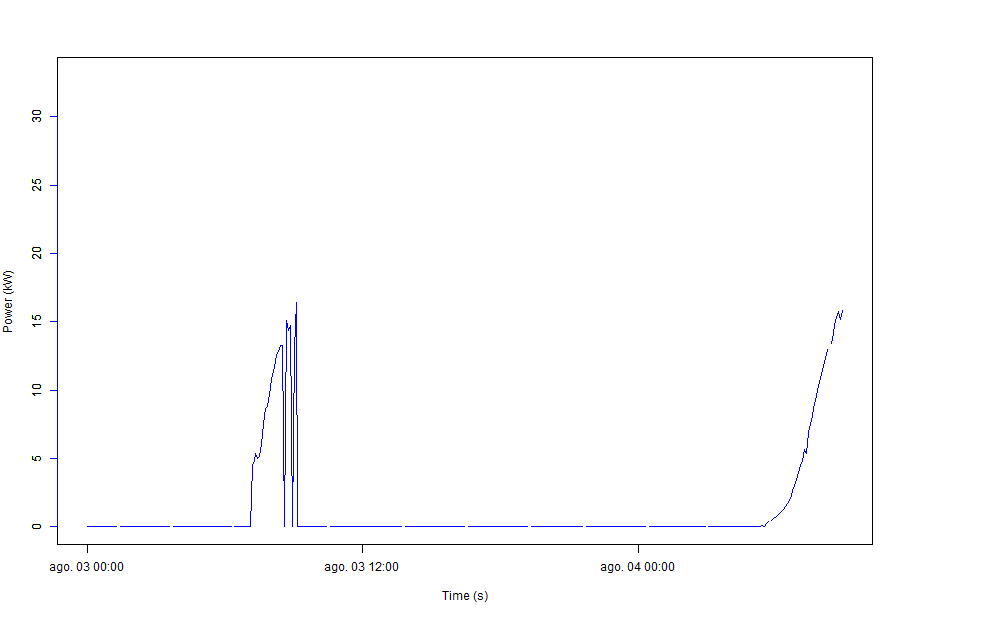}
\caption{Communication error.}
\label{error2}
 \end{subfigure}
 \caption{Power output (kW) of a solar inverter during periods affected by communication errors.}
 \label{error}
\end{figure}

In addition to power curtailment, the dataset also contained anomalies resulting from data recording errors. Figure \ref{error} shows two representative cases of these anomalies. They correspond to periods when the inverter was operating, but the data was incorrectly recorded, producing constant values over extended intervals. In Fig. \ref{error1}, this anomaly appears as two unrealistically flat power levels, the first around 7 kW and the second around 3 kW. In Fig. \ref{error2}, the anomaly is observed as an abrupt drop to zero in the recorded power at approximately 9:00 AM, after which the signal remains fixed at that value for the rest of the day, despite subsequent evidence of inverter activity.

Both types of anomalies, power curtailment and communication errors, were included in the labelling process performed by the experts.

\subsection{Experimental Setup}
\label{exp_fotovol}

The first five months of data, corresponding to normal operation, were used for training the models. The subsequent four months were used as the test set to evaluate their performance.

Two configurations of the Vacuum Spiker algorithm were considered. In the first configuration (Configuration 1), no recurrent layer was used, and the forward connection $I \rightarrow R$ was set to promote a predominantly depressing behaviour during training. In the second configuration (Configuration 2), the forward connection was set to favour the prevalence of potentiation, while the recurrent connection $R \rightarrow R$ was set to favour the prevalence of depression. Configuration 2 was the one more frequently identified as optimal throughout the experimental pipeline described in Section \ref{experimental}, as shown in Section \ref{resultados}. In contrast, Configuration 1, which lacks a recurrent connection, may offer greater energy efficiency and reduced variability in energy consumption over time, as suggested by Equation \ref{vacuum_macs} in Section \ref{energia}. This property is desirable, as predictable energy consumption facilitates estimating when the battery powering the device running the Vacuum Spiker algorithm will need to be replaced, which is especially useful in unattended environments, such as is the case here. For both configurations, the initial domain $D$ was set to the training set domain, and the compact interval bounding the subsequent domains $D^*$ was established as $I=[-10,170]$ kW. The remaining specific parameters used for both configurations are provided in Table \ref{parametros_solar}.
 
Accordingly, two versions of the Vacuum Spiker algorithm, each corresponding to one of the described parameter configurations, were trained and subsequently evaluated to assess their respective performance.

\begin{table}
 \footnotesize
 \centering
 \caption{Parameter configurations used in the implementation of the Vacuum Spiker algorithm for anomaly detection in solar inverters.}
 \label{parametros_solar}
 \begin{tabular}{|c|c|c|}
\toprule
\textbf{Parameter} & \textbf{Configuration 1} & \textbf{Configuration 2}\\ 
\midrule
$(A_-,A_+)$ for $I \rightarrow R$ & $(-0.1,-0.1)$ & $(0.1,0.1)$ \\
$(A_-,A_+)$ for $R \rightarrow R$ & Not used & $(-0.1,-0.1)$ \\
Resolution & $1$ kW & $1$ kW \\
Num. of neurons in $R$ & $1000$ & $1000$ \\
Spike threshold & $-55$ mV & $-55$ mV \\
$g_L$ & $1-e^{-1/100}$ & $1-e^{-1/100}$ \\
Interval size & $1$ kW & $1$ kW \\
Neurons resting potential & $-65$ mV & $-65$ mV\\ 
Neurons reset potential & $-65$ mV & $-65$ mV\\ 
Neurons refractory period & $5$ ms & $5$ ms\\
STDP $\tau_+$ & $1.051$ ms & $1.051$ ms\\
STDP $\tau_-$ & $1.051$ ms & $1.051$ ms\\
Neurons' $C$ constant & $1$ $\mu F$ & $1$ $\mu F$\\
\bottomrule
 \end{tabular}
\end{table}

The same evaluation metrics described in Subsection \ref{metricas} were used to compare the performance of both configurations. The evaluation procedure followed a similar approach to that described in Subsection \ref{procedimiento}. Specifically, for the G-Mean and F1-score metrics, multiple thresholds were applied to the spike-count time series to assess the potential occurrence of anomalies. The threshold that yielded the highest performance was selected. Ten threshold values were uniformly distributed between the minimum and maximum spike count values. Prior to performance evaluation, spike counts were optionally smoothed using a moving average over the previous $100$, $200$, or $300$ records. The best result obtained across all smoothing windows, including the one without smoothing, was retained.

\subsection{Results and Discussion}

The results obtained from evaluating the two configurations proposed in Section \ref{exp_fotovol} are presented in Table \ref{performance_energy_fotovol}. In that table, performance metrics, and the mean number of MAC operations required to perform inference on a single sample, are shown. The number of MAC operations was calculated following the methodology described in Section \ref{energia}. 

Configuration 2 corresponds to the combination of synaptic behaviours that most frequently produced the highest-performing models throughout the experiments described in Section \ref{experimental}, as shown in Table \ref{chi_tabla}, in Section \ref{resultados}. However, in the present case, Configuration 1 outperformed Configuration 2 across all metrics, and also demonstrated higher energy efficiency. It required less than half the number of MAC operations needed by Configuration 2, effectively doubling battery life. Moreover, the number of MAC operations required by Configuration 1 remained constant, making the energy consumption of this configuration more predictable over time. For these reasons, Configuration 1 was selected for anomaly detection in the solar inverter.

\begin{table}
 \footnotesize
 \centering
 \caption{
 Performance and energy consumption of Vacuum Spiker configurations for solar inverter anomaly detection. Each configuration is evaluated using G-Mean, F1-score, and AUC as evaluation metrics, while also reporting the average number of MAC operations required to perform inference on a single sample. The configuration achieving the best overall performance across the metrics and the lowest energy consumption is highlighted in bold.
 }
 \label{performance_energy_fotovol}
 \begin{tabular}{|c|c|c|c|c|}
\toprule
\textbf{Configuration} & \textbf{G-Mean} & \textbf{F1-score} & \textbf{AUC} & \textbf{MACs}\\
\midrule
\textbf{Configuration 1} & \textbf{0.6676} & \textbf{0.3185} & \textbf{0.7069} & \textbf{2000.0000} \\
Configuration 2 & 0.6569 & 0.2991 & 0.6763 & 5186.3950 \\
\bottomrule
\end{tabular}
\end{table}

\begin{figure}[H]
\centering
\subcaptionbox{Communication error. \label{comunicacion}}{
\includegraphics[width=0.47\textwidth]{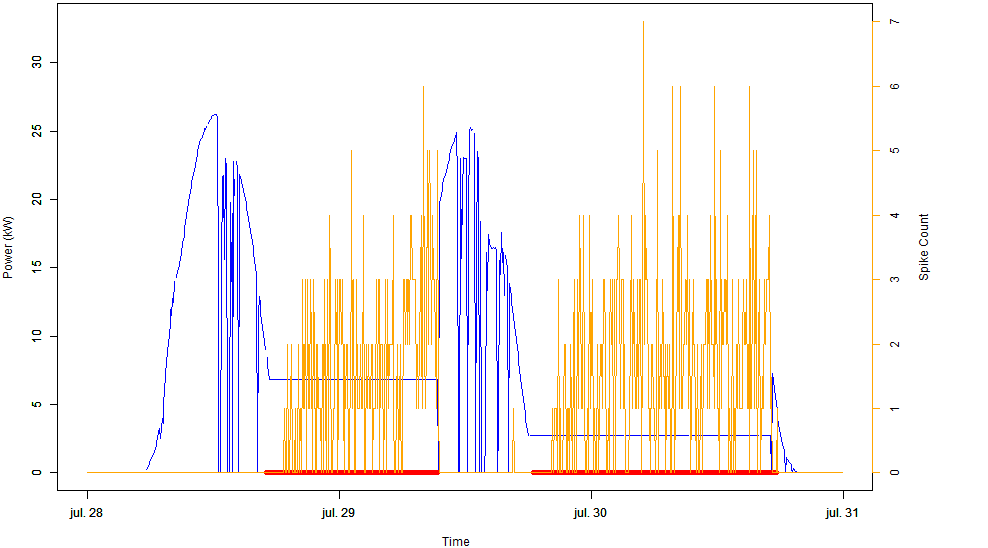}}
\subcaptionbox{Normal behaviour. Early June. \label{junio1}}{
\includegraphics[width=0.47\textwidth]{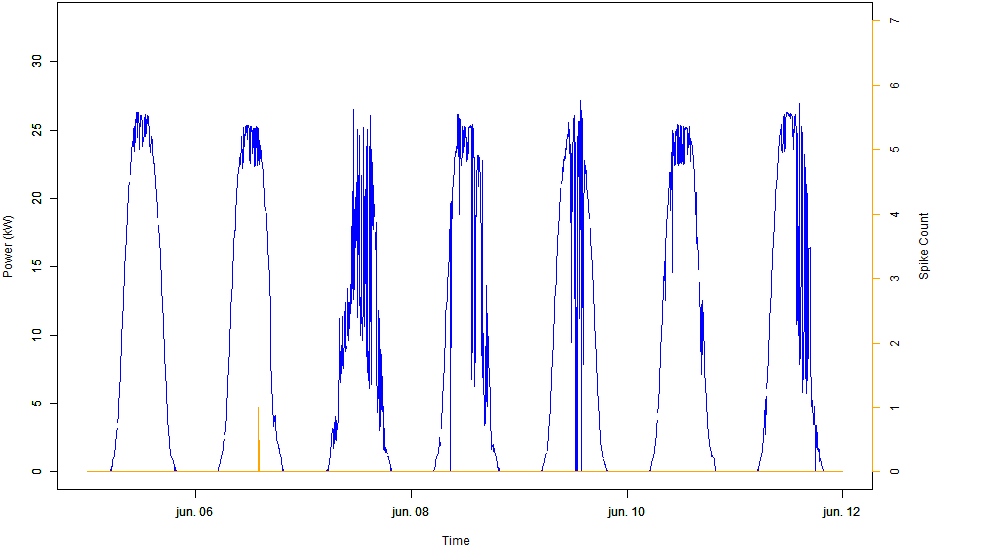}}
\subcaptionbox{Power curtailment. Late June. \label{junio2}}{
\includegraphics[width=0.47\textwidth]{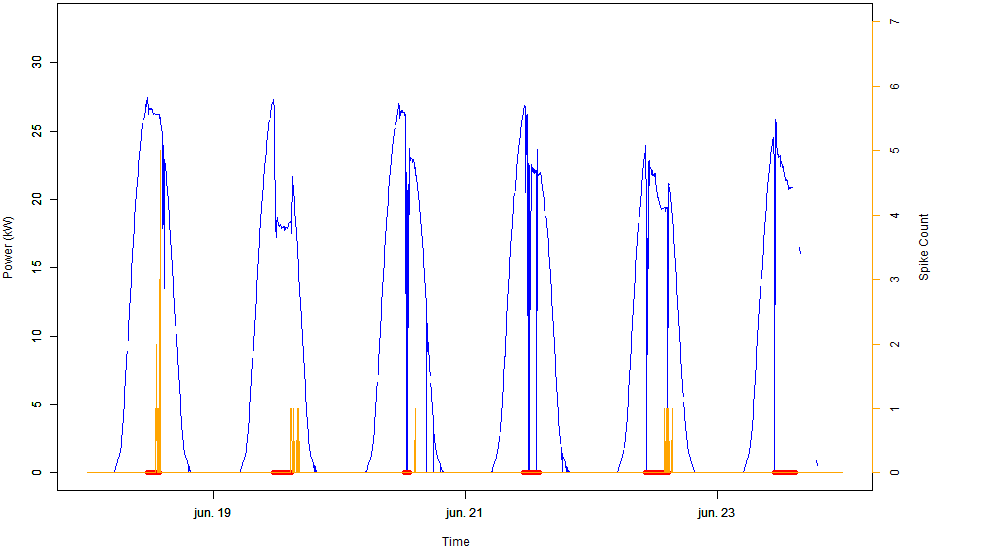}}
\subcaptionbox{Power curtailment. Early July. \label{julio0}}{
\includegraphics[width=0.47\textwidth]{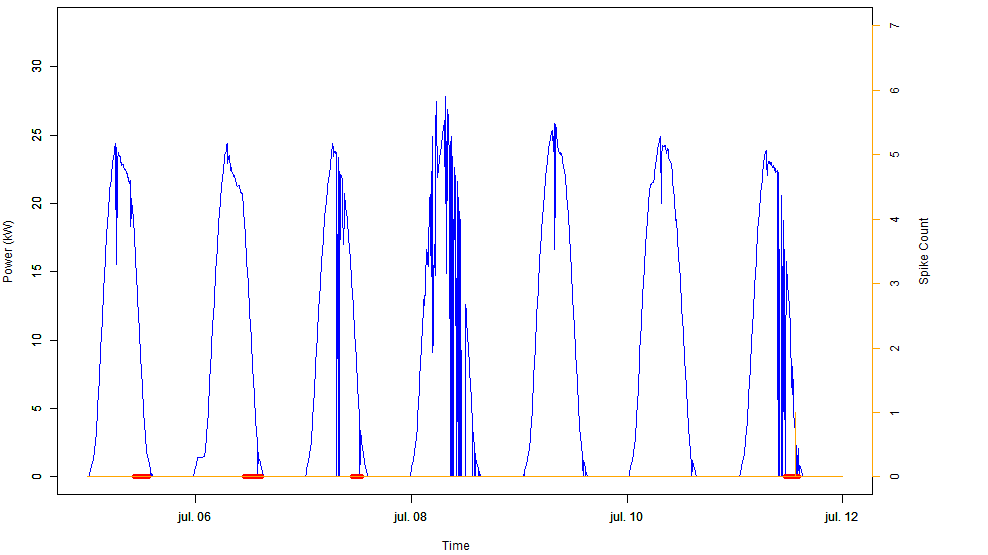}}
\subcaptionbox{Power curtailment. Middle July. \label{julio1}}{
\includegraphics[width=0.47\textwidth]{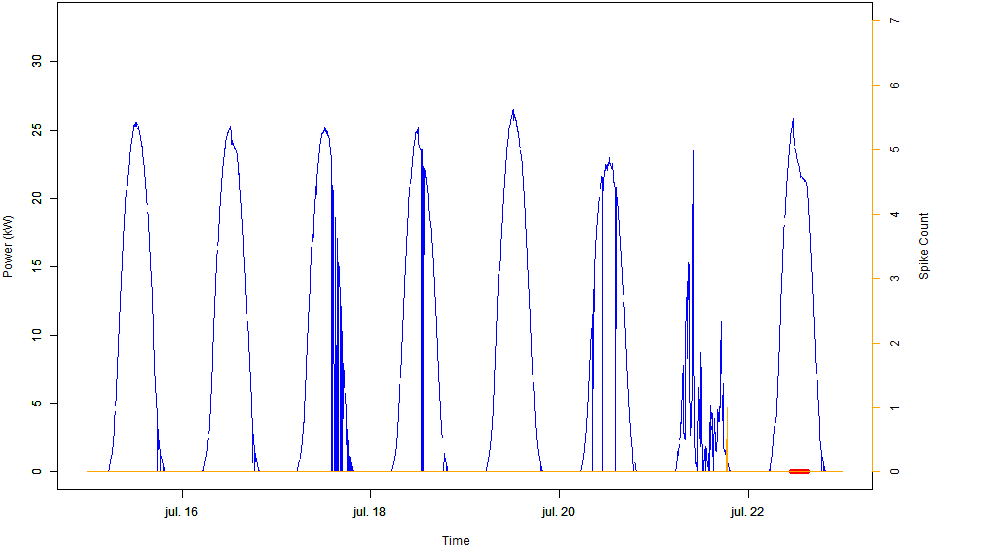}}
\subcaptionbox{Power curtailment. Late July. \label{julio2}}{
\includegraphics[width=0.47\textwidth]{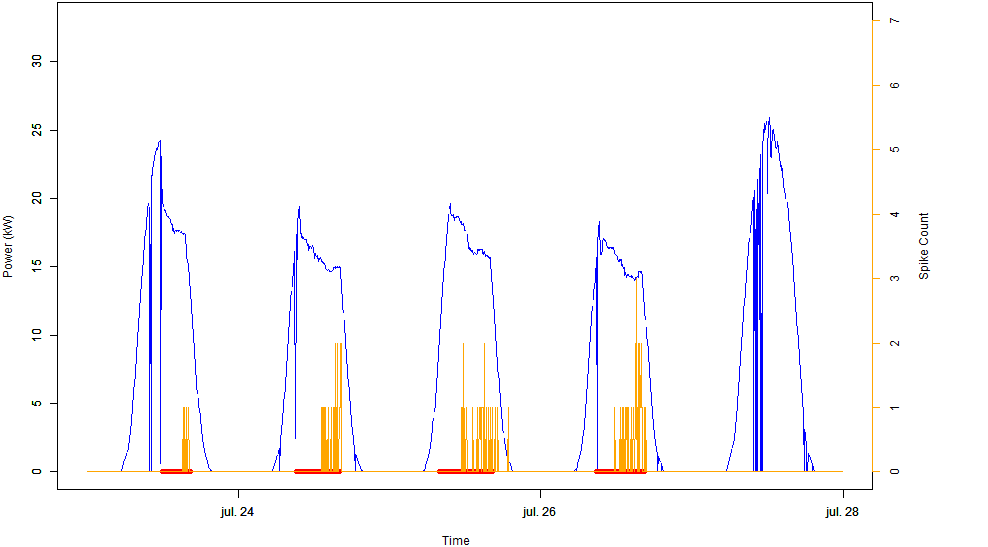}}
\subcaptionbox{Power curtailment. August. \label{agosto}}{
\includegraphics[width=0.47\textwidth]{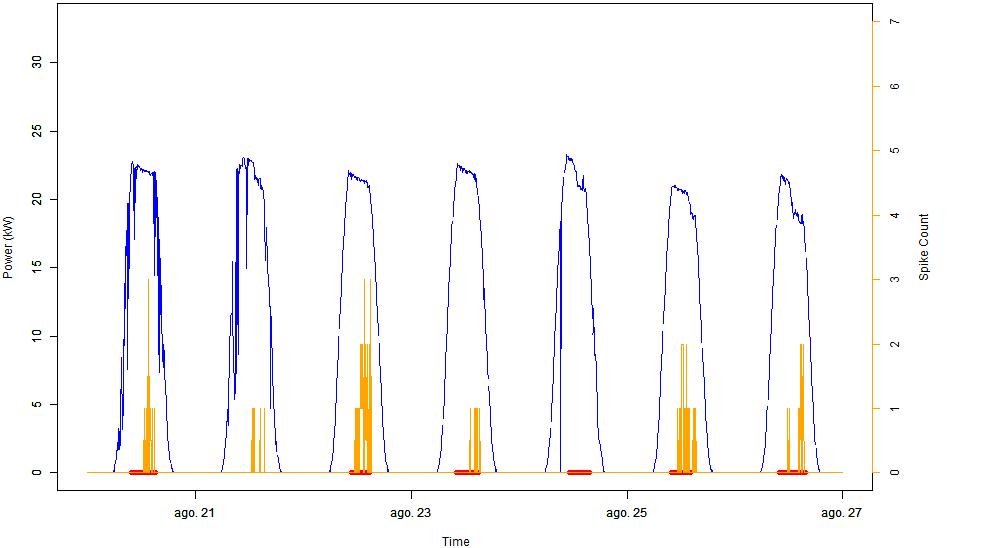}}
\subcaptionbox{Power curtailment. Late September. \label{septiembre}}{
\includegraphics[width=0.47\textwidth]{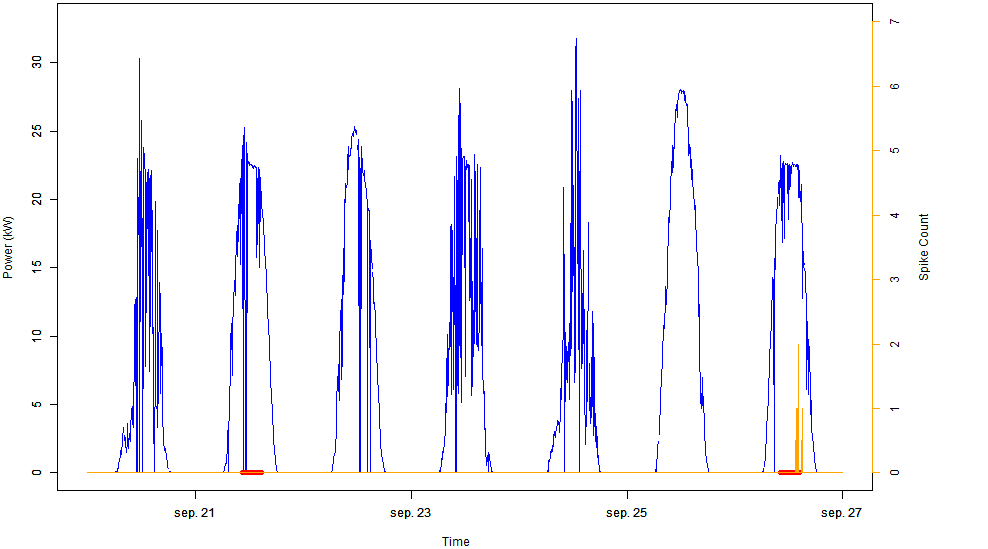}}
\caption{Power production (in kW) and anomaly detection for the studied solar inverter under different operational conditions. In all figures, the power output is shown in blue and corresponds to the left axis, while the number of spikes generated over time is shown in orange and corresponds to the right axis. Spikes are considered alerts. The scenarios depicted include normal operation, communication errors, and varying degrees of power curtailment across different months.}
\label{fig:solar_spikes}
\end{figure}

To illustrate the behaviour of Configuration 1 when processing data, Figure \ref{fig:solar_spikes} is presented. In its subfigures, the blue lines correspond to the power generated by the inverter, measured in kW. Their values are associated with the blue Y-axis on the left. Meanwhile, the orange lines, corresponding to the orange Y-axis on the right, indicate the number of spikes generated over time as the data is processed. Ground-truth anomalies are indicated by the red horizontal lines on the X-axis.

Under Configuration 1, the Vacuum Spiker algorithm tended not to generate spikes when processing data that exhibit normal behaviour, with only occasional ones appearing, as it can be observed in Figures \ref{junio1}, \ref{julio1}, and \ref{septiembre}. Since such spikes occurred infrequently, we considered the presence of one or more spikes as an alert, given that the number of false positives would remain low under this criterion. It is interesting to note that the algorithm did not emit spikes on days without anomalies but on which power production was notably irregular, likely due to meteorological conditions.

However, the spiking activity of the algorithm was higher when anomalies occurred. For instance, Figure \ref{comunicacion} shows a marked increase in spiking activity from the Vacuum Spiker during a communication error, resulting in a clear alert on that day.

Regarding power curtailment, until June 14th, when no clear power limitation was observed at the inverter, only one alert was generated. Figure \ref{junio1} illustrates some of those days. In contrast, during the remainder of the month, power curtailment became more apparent in the graphical representations. It was detected on eight days, resulting in alerts being triggered on five of them. Figure \ref{junio2} shows examples of days with clearer power curtailment, including three on which alerts were generated.

Subsequently, only two alerts were generated until July 11th, on days with power curtailment. During this period, five additional days exhibited power limitations, but no further alerts were triggered. Power limitations did not occur in the following days, and no alerts were generated until July 22. Two segments from these periods are shown in Figures \ref{julio0} and \ref{julio1}. However, starting on July 23rd, power limitations began to happen almost every day. Between July 22nd and July 31st, the Vacuum Spiker algorithm triggered alarms on six days. In August, the number of days with alerts increased to $24$, with power limitations occurring on most of days. Figures \ref{julio2} and \ref{agosto} illustrate several of these days, in which power curtailment was clearly visible.

In the first half of September, alerts were generated on seven days, while power curtailment occurred on ten days, following a pattern similar to that observed in August. However, only two alerts were generated in the second half of the month, one of them corresponding to a day when power curtailment occurred. As shown in Figure \ref{septiembre}, power curtailment was labelled on only two days during this period. Meteorological conditions make it difficult to assess whether it may have occurred on additional days. In contrast, several days within this period exhibited normal inverter behaviour. Figure \ref{septiembre} presents two of these days, along with the corresponding spikes.

In conclusion, the proposed Vacuum Spiker algorithm appears to be effective in detecting power curtailment in the studied inverter. Spiking activity under normal operating conditions was found to be very low, in contrast to the behaviour observed during curtailment. When at least one spike is considered an alert, such alerts were generated frequently from the onset of power limitation. As shown in Figures \ref{fig:solar_spikes}, alerts seemed to occur more often when power curtailment was more pronounced. Also, a delay tends to occur between the onset of the anomaly and the response of the Vacuum Spiker algorithm. Although this may affect the performance metrics presented in Table \ref{performance_energy_fotovol}, it is unlikely to be relevant for the present use case, since replacing a damaged inverter typically requires several days, and therefore detection within minutes or hours is not critical. Nevertheless, our system could potentially have enabled the replacement of the damaged inverter few time after the first signs of malfunction appeared. This would likely have increased the plant’s energy production during the study period.

It is worth noting that the developed model was able to remain unresponsive to noise introduced by adverse meteorological conditions, as illustrated in Figure \ref{fig:solar_spikes}. This indicates that the Vacuum Spiker algorithm was robust in distinguishing patterns and shapes caused by rain, clouds, etc., from those resulting from inverter malfunctions or communication issues. Furthermore, its energy consumption was remarkably low, as shown in Table \ref{performance_energy_fotovol}. Therefore, the Vacuum Spiker algorithm has demonstrated its potential as an ideal candidate for performing anomaly detection in solar inverters under strict energy constraints.

\FloatBarrier

\section{Conclusion and Future Work}
\label{conclusion}

In this paper, the Vacuum Spiker algorithm is proposed, which performs anomaly detection on time series data. This is accomplished by monitoring neuronal activity in the hidden layer. A novel coding scheme is employed, requiring a single spike per input sample, transmitted within just one time step. The STDP learning rule is adapted to maintain low spiking activity in the hidden layer under normal conditions, while it is increased when an anomaly occurs. These design features are intended to enhance the Vacuum Spiker algorithm energy efficiency.

Through extensive empirical evaluations on a diverse set of publicly available time series datasets, the Vacuum Spiker algorithm has demonstrated performance on par with to the best-performing deep learning-based anomaly detection models. The only more energy-efficient model, LOF, a traditional machine learning approach, achieved significantly lower performance. However, Vacuum Spiker attains its results while consuming several orders of magnitude less energy than the rest of the evaluated algorithms, highlighting its potential as a viable alternative in scenarios where computational or energy resources are severely limited, such as edge computing environments, embedded systems, and wearable devices. This is further exemplified by a real-world application, which shows its effectiveness in facilitating the detection and subsequent replacement of damaged devices in an unattended industrial setting.

This study lays the groundwork for further research into the use of SNNs for anomaly detection tasks. Future investigations could explore variations in network architecture, or experiment with different synaptic plasticity rules. Such developments may yield further improvements in both accuracy and efficiency, reinforcing the relevance of SNN-based models in the broader context of time series anomaly detection. On the other hand, the way Interval Coding operates could facilitate the application of SNN models to online learning scenarios, where rapid adaptation to changing patterns may be critical.

\FloatBarrier
\section*{Acknowledgemetns}

This work was supported by the Ministry of Science and Innovation with project PID2023-149511OB-I00, and under the programme for mobility stays at foreign higher education and research institutions "José Castillejo Junior" with code CAS23/00340, and by the CDTI project (CER-20231019 (CICERO) and ICECyL project (Junta de Castilla y León) under project CCTT5/23/BU/0002 (QUANTUMCRIP).

\section*{Declaration of Generative AI and AI-assisted Technologies in the Writing Process}

During the preparation of this work, the authors used ChatGPT-5 in order to 
improve the readability and language of the manuscript. After using this service, the authors reviewed and edited the content as needed and take full responsibility for the content of the published article.

\section*{Competing Interests}
The authors declare that they have no competing interests.

%
%
%

\bibliographystyle{plainnat}
\bibliography{biblio}

\appendix
\section{Estimating Required MACs in Baselines}
\label{apendice}
\subsection{Traditional ANN Models}

Respect to traditional ANN models used in this study as baselines, they are composed mainly by three types of layers: convolutional layers, LSTMs, and dense layers. Moreover, they also contain average pooling and batch normalization operations, which also require MAC operations. The equations to estimate the number of MAC operations for those kinds of layer and operations are the following:
\begin{itemize}
\item \textbf{\textit{Dense Layers}}

As a dense layer can be considered as a matrix multiplication combined with a non-linear activation, the number of MAC operations $M_D$ required to apply it to a single sample can be estimated following Eq. \ref{densa}:
\begin{equation}
M_D=N_I N_O
\label{densa}
\end{equation}
\noindent where $N_I$ is the number of input neurons, and $N_O$ corresponds to the number of output neurons.

\item \textbf{\textit{Convolutional Layers}}

For each output neuron in this kind of layers, the kernel has to be convolved with the input channels. As the output size is repeated along each output channel, the number of MAC operations $M_C$ required by an inference on a convolutional layer can be estimated as described in \ref{convolucional}.
\begin{equation}
M_C=K C_I C_O O
\label{convolucional}
\end{equation}
\noindent where $C_I$ is the number of input channels, $C_O$, the number of output channels, $O$, the output size and $K$ the kernel size.

\item \textbf{\textit{LSTMs}}

LSTMs show a more complex structure. The procedure for processing a single sample can be found in \citep{DBLP:journals/corr/SakSB14}. By following that procedure, it is possible to arrive to Eq. \ref{lstm} to count the number $M_{LSTM}$ required to compute the MAC operations required to process a single sample on a LSTM layer:
\begin{equation}
M_{LSTM}= T (4 n c + 4 n ^2 + 12 n)
\label{lstm}
\end{equation}
\noindent where $T$ is the length of the input sequence, $n$, the number of neurons, and $c$, the number of input features.

\item \textbf{\textit{Batch Normalization Operators}}

Batch normalization, in inference, normalizes the inputs by subtracting the mean $\mu$, and by dividing them by the standard 
deviation, $\sigma$, both of which are computed during training. This operation can be expressed as in \ref{batch}.
\begin{equation}
o=\frac{i}{\sigma}-\frac{\mu}{\sigma}
\label{batch}
\end{equation}
\noindent where $i$ and $o$ denote the input and output, respectively. If the ratio $-\mu/\sigma$ is saved during training, batch normalization during inference requires only one multiplication and one adition per input element. Therefore, the number of MAC operations is equal to the size of the data being processed.

\item \textbf{\textit{Average Pooling}}

Average pooling computes the mean of the input values within each region defined by a kernel of predetermined size as it is applied across the input tensor. Therefore, computing each output value requires $K-1$ additions and one multiplication, where $K$ is the kernel size. This results in $K$ MAC operations per output value.

\end{itemize}

\subsection{Machine Learning Models}

The number of MAC operations required to perform inference with the two machine learning algorithms used as baselines can be estimated as follows:
\begin{itemize}
\item \textbf{\textit{OCSVM}}

For an OCSVM with an RBF kernel, the inference operation is expressed as in Eq. \ref{ocsvm}:
\begin{equation}
OCSVM(x) = \sum_{i=1}^n \alpha_i e^{-\gamma ||x-x_i||^2}
\label{ocsvm}
\end{equation}
\noindent where $x_i$ denotes the $i$-th support vector, $\alpha_i$ are constants associated with each support vector, and $\gamma$ is a scalar constant. $n$ represents the number of support vectors. High values of $OCSVM(x)$ indicate anomalies, whereas lower values correspond to normal data points.

Computing the squared distance term $||x - x_i||^2$ requires $2d$ MAC operations, where $d$ is the dimensionality of the vectors. Two additional MAC operations are needed for each support vector due to the multiplications by $\alpha_i$ and $-\gamma$. Since this process is repeated for each of the $n$ support vectors, and the results for each of them can be accumulated during execution, the total number of MAC operations required to compute a single sample using an OCSVM, $M_{OCSVM}$, can be estimated as shown in \ref{mac_ocsvm}:
\begin{equation}
M_{OCSVM}=n(2d+2)
\label{mac_ocsvm}
\end{equation}

\item \textbf{\textit{LOF}}

To perform inference using the LOF algorithm, the Euclidean distance between the input vector $x$ and each of the $n$ training vectors must be computed. As in the previous case, this requires $2d$ operations per training vector, where $d$ denotes the vector dimensionality. Subsequently, the reachable distance is computed for each point; however, this step does not involve any MAC operations. Next, the inverse local density and the final LOF value are calculated. The inverse local density requires computing a mean and an inverse, for the $k$ nearest neighbours of a point, which entails $k+1$ operations. Since this computation must be applied to each of the $k$ neighbours and the input point $x$, the total number of MAC operations involved is $(k+1)^2$. On the other hand, the final calculation of the LOF value requires an additional number of $k+1$ operations. Therefore, the total number of MAC operations required by the LOF algorithm, $M_{LOF}$, can be estimated as it can be seen in \ref{lof}:
\begin{equation}
M_{LOF} = 2d+(k+1)^2+k+1
\label{lof}
\end{equation}
\end{itemize}

\end{document}